\documentclass[journal]{IEEEtran}
\usepackage{amsmath}  % 支持equation*、\text{}、数学符号，必加！
\usepackage{algorithm}  % 提供algorithm浮动体环境，必加！

\usepackage{algpseudocode}  % 提供algorithmic环境和\Require\For\State等命令，必加！
\usepackage[table]{xcolor}
\definecolor{baselinerow}{gray}{0.92}
\usepackage{amsfonts,amssymb}
\usepackage{array}
\usepackage[caption=false,font=normalsize,labelfont=sf,textfont=sf]{subfig}
\usepackage{textcomp}
\usepackage{stfloats}
\usepackage{url}
\usepackage{threeparttable}
\usepackage{verbatim}
\usepackage{graphicx}
\usepackage{booktabs}
\usepackage{multirow}
\usepackage{makecell}
\usepackage{pgfplots}
\usepackage{xcolor}  
\usepackage{graphicx}
\usepackage[colorlinks=true, linkcolor=blue, citecolor=blue, urlcolor=black]{hyperref}

\graphicspath{
    {../Latex/Figures/}        % 主文件 TMM.tex 的路径
    {../../Latex/Figures/}     % Sections/*/*.tex 路径
    {../../../Latex/Figures/}  % 如果子文件深度更深可以继续加
}

\pgfplotsset{compat=1.18}

\def\BibTeX{{\rm B\kern-.05em{\sc i\kern-.025em b}\kern-.08em
    T\kern-.1667em\lower.7ex\hbox{E}\kern-.125emX}}
\usepackage{balance}
\begin{document}
\title{GIRL-DETR: Gradient-Isolated Reinforcement Learning for Video Moment Retrieval}

\author{Shihang Zhang, Mingjin Kuai, Ye Wei, Zhen Zhang, and Wei Ji
\thanks{Shihang Zhang and Mingjin Kuai contributed equally to this work.}
\thanks{Corresponding author: Wei Ji.}
\thanks{Shihang Zhang and Ye Wei are with the College of Electronics and Information Engineering, Sichuan University, Chengdu 610065, China   (e-mails:                   \url{shihang_zhang@stu.scu.edu.cn},
\url{weiye_@stu.scu.edu.cn}).}
\thanks{Mingjin Kuai, Zhen Zhang, and Wei Ji are with the School of Intelligence Science and Technology, Nanjing University, Suzhou 215163, China   (e-mails:                \url{mingjinkuai@smail.nju.edu.cn}, \url{zhen_zhang@nju.edu.cn}, \url{weiji@nju.edu.cn}).}}
\markboth{IEEE TRANSACTIONS ON IMAGE PROCESSING}%
{How to Use the IEEEtran \LaTeX \ Templates}

\maketitle

\begin{abstract}
Video Moment Retrieval (VMR) task requires accurately localizing temporal boundaries aligned with natural language queries, but many models suffer from a misalignment between continuous surrogate losses and non-differentiable metrics, leading to optimization stagnation during the late stages of training and trapping boundary predictions in suboptimal solutions. Although Reinforcement Learning (RL) post-training successfully optimizes localization results for large models, applying it directly to lightweight networks easily disrupts the fragile feature representations established during the supervised phase. To overcome this optimization bottleneck, we propose \textbf{G}radient-\textbf{I}solated \textbf{R}einforcement \textbf{L}earning for \textbf{DETR} (GIRL-DETR), introducing RL post-training into a lightweight temporal localization framework for the first time. The input video and text features first establish early alignment through Cross-Modal Interaction (CMI) before entering the transformer encoder. Subsequently, a Text-Guided Gating (TGG) mechanism dynamically injects semantic priors into the queries before the transformer decoder generates candidate proposals, providing high signal-to-noise ratio inputs for temporal prediction. After the supervised training reaches convergence, the backbone network is frozen to protect the feature manifold, while the detection head directly optimizes the non-differentiable evaluation metric tIoU to enhance localization accuracy through a Three-stage Progressive Reinforcement Learning (TPRL) strategy. This approach achieves an orthogonal decoupling of state representation and metric optimization. Experiments on Charades-STA, QVHighlights, and TACoS demonstrate that GIRL-DETR effectively resolves surrogate loss degradation and achieves substantial accuracy improvements with minimal parameter updates, providing a robust new pathway for RL applications in lightweight VMR models. The code is available at \url{https://github.com/Z-Shihang/GIRL-DETR}

\end{abstract}
\section{Introduction}

\IEEEPARstart{W}{ith} the rapid growth of video data, video moment retrieval (VMR), which localizes target segments based on natural-language queries, has emerged as a fundamental task in video understanding~\cite{gao2017tall,anne2017localizing,sun2021maban,zhang2023temporal,hu2021video}. Compared to coarse-grained retrieval, VMR imposes stricter requirements on boundary precision, thus necessitating both deep cross-modal semantic alignment and accurate temporal boundary regression~\cite{yang2021local,yang2022video}. However, in practical localization scenarios, the predicted segments frequently exhibit loose boundaries or global temporal shifts~\cite{lee2024bam,seol2023bmrn,sun2021maban}. These issues directly limit the upper bound of the temporal Intersection-over-Union (tIoU) and degrade the ranking quality of high-confidence candidate segments. Therefore, developing an efficient framework that preserves accurate cross-modal semantic alignment while strictly supporting fine-grained temporal boundary regression and the reliable ranking of high-confidence candidates remains a core challenge.

\begin{figure}[t]
  \centering
  \includegraphics[width=\columnwidth]{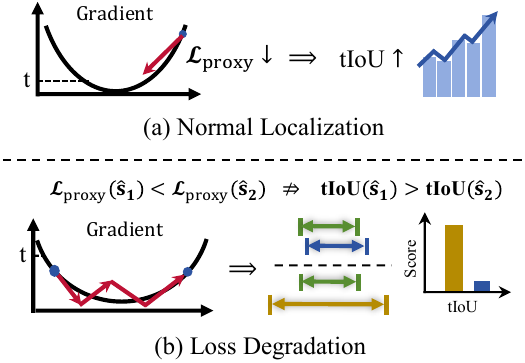}
  \caption{\textbf{Loss Degradation and Ranking Collapse.} (a) In early training, proxy loss optimization effectively drives metric improvements, despite gradient updates deviating from the optimal path. (b) Near the optimal solution, loss degradation prevents the model from finding effective update directions aligned with the localization metric. This misalignment triggers ranking collapse, erroneously prioritizing suboptimal predictions.}
  \label{fig:mismatch}
\end{figure}

To address this challenge, existing studies primarily follow two paradigms: methods based on multimodal large models~\cite{wang2026time,chen2025datasets} and lightweight models designed for efficient inference~\cite{lei2021detecting,Moon2025CorrelationguidedCO,sun2024tr,yang2022video,hu2021video}. Despite the architectural differences, both paradigms predominantly rely on local proxy losses for supervised training~\cite{Moon2025CorrelationguidedCO,liu2022umt,xiao2024bridging,yang2021local}. This reliance commonly leads to a bottleneck in temporal alignment during the late stages of optimization. As illustrated in Fig.~\ref{fig:mismatch}, this phenomenon stems from an inherent discrepancy between the optimization objective and the evaluation metric, referred to as the loss-metric mismatch. The geometric constraints imposed by local proxy losses fail to fully reflect the global alignment quality of the tIoU. Consequently, this discrepancy renders models prone to suboptimal states, wherein the performance of localization and ranking stagnates or degrades toward the end of training.

To bridge this mismatch, specific multimodal large models introduce Reinforcement Learning (RL) during the fine-tuning phase of targeted components to directly optimize global evaluation metrics~\cite{ouyang2022training,sun2024aligning,feng2026video}. However, when rewards and penalties derived from the tIoU are introduced into lightweight VMR architectures~\cite{hu2021video}, the high-variance gradients induced by policy exploration and sampling cause drastic parameter updates. These updates disrupt the multimodal feature distributions established during supervised training, which results in the degradation of representations~\cite{williams1992simple,sutton1999policy,kirkpatrick2017overcoming,li2017learning}. Consequently, the stable application of RL in lightweight networks for the direct optimization of evaluation metrics remains underexplored.

To resolve these issues, we propose GIRL-DETR, a framework based on the DETR architecture that incorporates a text-guided gating mechanism and employs a gradient-isolated three-stage RL post-training strategy. After the proposed base cross-modal fusion network achieves convergence under supervision, GIRL-DETR applies a gradient isolation strategy to directly calibrate the temporal localization. This strategy strictly freezes the modules for feature fusion and alignment to block the backpropagation of high-variance RL gradients, while permitting reward-driven exploration guided by the tIoU solely within the lightweight prediction head. This design prevents the degradation of representations at the architectural level, achieves an orthogonal decoupling between the multimodal state representations and the reward mechanism of the tIoU, and enhances the distributions of both temporal boundaries and confidence scores.

Overall, our main contributions are summarized as follows:

\begin{itemize}

    \item We propose GIRL-DETR to address the universal loss-metric mismatch in VMR. It first establishes stable multimodal representations via supervised training, then directly optimizes localization through RL post-training, achieving robust performance across Charades-STA, QVHighlights, and TACoS.

    \item To overcome task-agnostic queries, we introduce a Text-Guided Gating (TGG) mechanism. It dynamically injects global text priors into initial queries, providing high signal-to-noise ratio inputs that significantly enhance semantic awareness for temporal decoding.

    \item Traditional proxy losses are inherently misaligned with the non-differentiable tIoU. We are the first to introduce RL post-training into lightweight VMR models to achieve direct metric optimization, utilizing a gradient isolation strategy that protects backbones with negligible overhead.

    \item We design a Three-stage Progressive Reinforcement Learning (TPRL) strategy to ensure stable optimization. Extensive experiments demonstrate that TPRL is a highly generalizable post-training strategy, effectively adaptable to representative baselines and feature backbones.

\end{itemize}
\section{Related Work}

\subsection{Video Moment Retrieval}
VMR aims to localize temporal segments relevant to queries within untrimmed videos~\cite{gao2017tall,anne2017localizing,liu2021context}. Early approaches primarily focus on proposal generation, cross-modal matching, dense temporal modeling, and boundary regression~\cite{regneri2013grounding,yuan2019find,zhang2020learning,zhang2020span,zeng2020dense,chen2020learning,zhang2021multi}. Recently, the field of VMR has advanced along two primary trajectories: generative methods based on large models and lightweight discriminative models. Generative approaches utilizing large models, including Time-R1~\cite{wang2026time} and TVG-R1~\cite{chen2025datasets}, formulate temporal localization as generative reasoning. These methods benefit from robust semantic understanding but incur substantial computational costs. Within the category of lightweight models, online methods such as GTR~\cite{cao2021pursuit} update visual representations end-to-end for tighter adaptation to text queries. In contrast, offline methods, including Moment-DETR~\cite{lei2021detecting} and QD-DETR~\cite{moon2023query}, rely on pre-extracted video features for high inference efficiency.

To establish a unified formulation on top of these frozen representations, DETR-based architectures have emerged as a prominent paradigm by treating temporal localization as direct set prediction. For instance, Moment-DETR~\cite{lei2021detecting} formulates moment retrieval as a dense prediction task. This formulation eliminates the requirement for hand-crafted anchor designs or heuristic post-processing techniques, thereby streamlining a fully differentiable prediction pipeline. Subsequently, methods such as QD-DETR, CG-DETR~\cite{Moon2025CorrelationguidedCO}, EaTR~\cite{jang2023knowing}, and BAM-DETR~\cite{lee2024bam} further enhance retrieval accuracy through query-dependent video representations, cross-modal calibration, event-aware query modeling, and boundary-aligned decoding, respectively. Building upon the DETR framework, and distinguishing our approach from existing methods relying on supervised proxy losses, we introduce RL for post-training into the decision head. This directly aligns the optimization process with the final evaluation metric of tIoU.

\subsection{Post-Training}

In recent years, post-training has emerged as a crucial paradigm to improve task adaptation and objective alignment~\cite{christiano2017deep,stiennon2020learning}. The core concept involves further calibrating model behavior through an additional fine-tuning or alignment stage after base training. For large language models, InstructGPT~\cite{ouyang2022training} initially applies supervised fine-tuning and subsequently performs RLHF; Llama 2-Chat~\cite{touvron2023llama} adopts a similar combination. Beyond RL-based alignment, DPO~\cite{rafailov2023direct} transforms preference learning into a direct optimization objective, eliminating the need for an explicit reward model. Furthermore, continual learning and parameter-efficient fine-tuning demonstrate stable adaptation during post-training~\cite{dettmers2023qlora}. EWC~\cite{kirkpatrick2017overcoming} constrains critical parameters to prevent catastrophic forgetting, whereas LoRA and adapter-based methods~\cite{hu2022lora,houlsby2019parameter} update only a limited number of newly introduced parameters to preserve existing capabilities. Building upon these paradigms for parameter efficiency and alignment, we adopt a progressive three-stage post-training strategy based on RL. By updating only lightweight modules while freezing the majority of parameters, we effectively decouple the learning of stable cross-modal representations from the final objective alignment.

\subsection{Reinforcement Learning for Temporal Localization}

RL has demonstrated substantial efficacy in optimizing non-differentiable decision objectives and was incorporated into early research on temporal localization. In this context, it primarily served to formulate policies for search operations or boundary adjustments~\cite{williams1992simple,caicedo2015active}. For instance, SM-RL~\cite{he2019read} formulates language-driven localization as a recurrent decision process and utilizes semantic rewards to guide stepwise searches. TSP-PRL~\cite{wu2020tree} designs a tree-structured progressive policy that continuously adjusts candidate boundaries through action decisions. These methods validate the effectiveness of RL in sequential boundary adjustment. However, they formulate the entire temporal localization process as a recurrent Markov Decision Process. Consequently, they do not explore utilizing RL as a post-training alignment strategy to address the inherent discrepancy between continuous proxy losses and non-differentiable evaluation metrics within modern dense prediction architectures. Recently, methods based on large models, including Time-R1~\cite{wang2026time}, have incorporated tIoU rewards for alignment. However, because these methods output tokenized timestamps, their optimization target remains the discrete probability of text generation. This target fundamentally differs from the continuous boundary regression required by lightweight detection models. As highlighted previously, the high gradient variance of RL makes lightweight networks susceptible to training instability, forcing existing methods to rely on proxy losses~\cite{lee2024bam,lei2021detecting}.

\begin{figure*}[t]
  \centering
  \includegraphics[width=\linewidth]{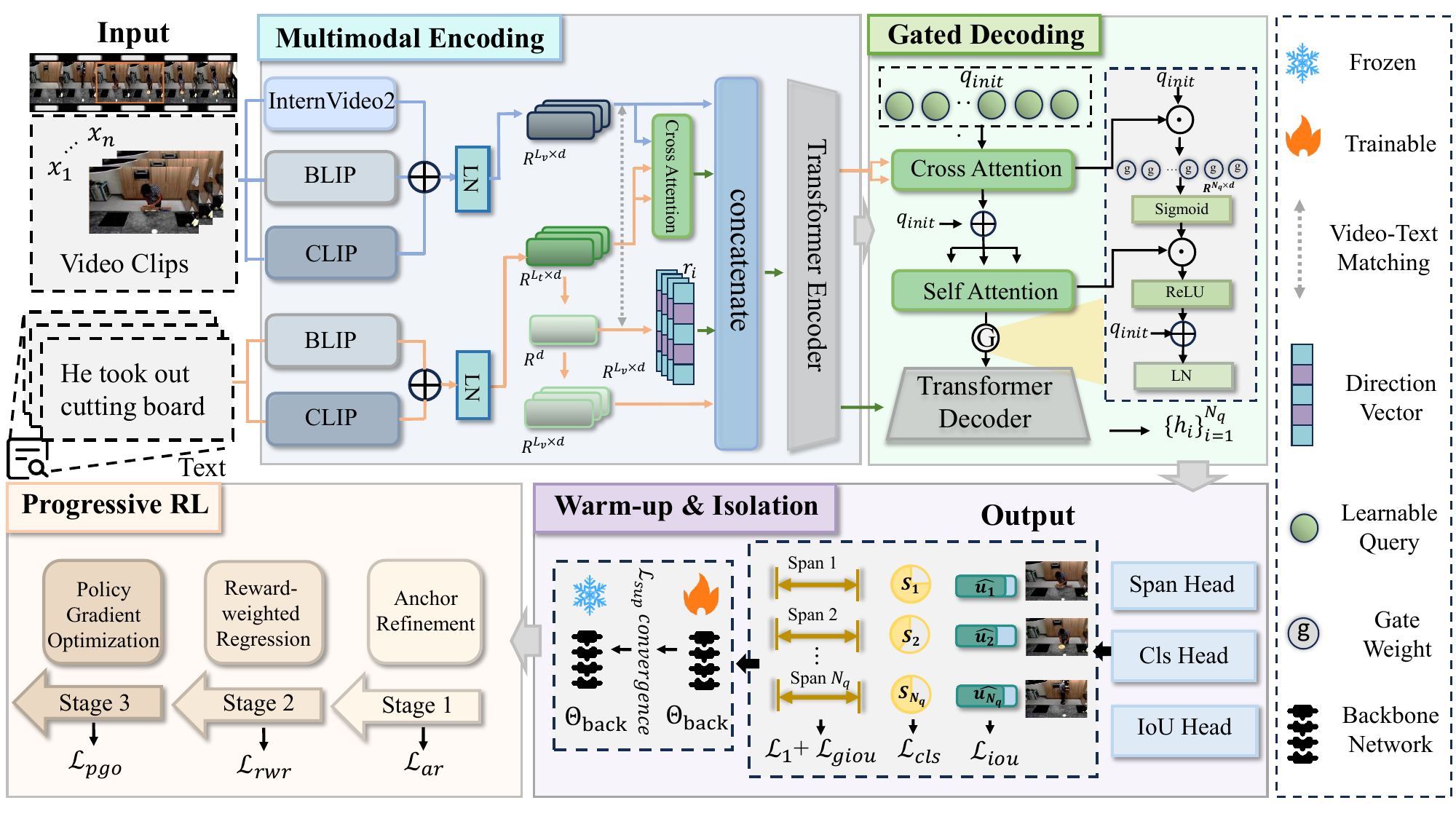}
  \caption{\textbf{ The overall architecture of the proposed GIRL-DETR framework.} The pipeline processes video and text inputs through sequential stages of multimodal encoding and semantic-gated decoding to generate segment candidates. After achieving supervised convergence, the backbone is frozen while the detection head is refined via progressive reinforcement learning to optimize the tIoU.}
  \label{fig:pipeline}
\end{figure*}

To the best of our knowledge, we are the first to introduce RL-based post-training into lightweight VMR models to directly optimize the non-differentiable evaluation metric. By adopting a gradient-isolated progressive strategy that freezes the backbone to protect fragile feature representations and updates only the detection head, our approach achieves an orthogonal decoupling of state representation and metric optimization. This decoupling effectively overcomes the optimization bottleneck caused by surrogate loss degradation.
\section{Method}

\subsection{GIRL-DETR Architecture}
Given an untrimmed video $\mathcal{V}$ and a natural language query $\mathcal{Q}$, the GIRL-DETR framework establishes a pipeline to predict target boundaries, as illustrated in Figure~\ref{fig:pipeline}. The architecture operates through four sequential stages. First, in the multimodal feature encoding stage, heterogeneous video and text feature streams are fused into a joint feature matrix within the encoder. Second, the guided gating and decoding stage injects semantic priors into the initial query vector $\mathbf{q}$, which the decoder subsequently utilizes to generate target segment candidates. Third, the supervised warm-up and gradient isolation stage trains the entire network using proxy losses. Once convergence is achieved, the optimization objective shifts, and the feature extraction backbone is frozen, leaving only the detection head active for training. Finally, in the progressive reinforcement learning stage, the detection head transitions smoothly from a supervised warm-up to policy gradient optimization. This progression introduces an RL mechanism to finalize the optimization of the localization.

\subsection{Text-Aware Multimodal Encoding}
\textit{  1) Independent Feature Extraction.} The video $\mathcal{V}$ is uniformly sampled into $L_v$ clips, which are independently processed by three distinct visual encoders. These pre-trained encoders produce clip descriptors with varying intrinsic dimensions. The three streams of descriptors are concatenated and mapped into a unified clip embedding $\mathbf{V} \in \mathbb{R}^{L_v \times d}$ via a linear projection. For the natural language query $\mathcal{Q}$, a parallel set of two pre-trained text encoders is utilized to extract linguistic features. These features are concatenated and projected to form the text representation $\mathbf{T} \in \mathbb{R}^{L_t \times d}$, which shares the same feature dimension $d$ with the video representation.

\textit{  2) Cross-Modal Interaction Module.} Standard cross-attention layers in the encoder struggle with alignment efficiency when simultaneously handling alignment and understanding tasks on raw visual features. To address this issue, a lightweight cross-modal interaction module is introduced before the encoder. This module injects text priors into each clip from three complementary perspectives, enabling the video features to acquire query awareness prior to encoding. First, attention pooling is applied to the text sequence $\mathbf{T}$ to derive a global sentence summary $\mathbf{s} \in \mathbb{R}^{d}$, which serves as a global semantic anchor. Second, the cosine similarity $r_i$ between the clip embedding $\mathbf{V}_i$ and the summary $\mathbf{s}$ is computed. This scalar scales a learnable direction vector $\mathbf{u} \in \mathbb{R}^{d}$ to produce a continuous correlation feature $r_i\mathbf{u}$. This operation ensures that highly relevant and irrelevant clips receive distinguishable modulation in the feature space. Third, the video features are utilized as queries to perform cross-attention on the text, retrieving word-level semantic context:
\begin{equation}
\mathbf{C} = \mathrm{Attention}(\mathbf{V}, \mathbf{T}, \mathbf{T}\mathbf{W}_v),
\end{equation}
where $\mathbf{W}_v \in \mathbb{R}^{d \times d}$ is a learnable value projection matrix that allows the model to adaptively aggregate text semantics. These three signals are concatenated with the original clip features and fused through a shared linear projection to obtain the text-aware video clip embedding:
\begin{equation}
\tilde{\mathbf{V}}_i = \mathrm{ReLU}(\mathbf{W}_f (\mathbf{V}_i \oplus (r_i\mathbf{u}) \oplus \mathbf{s} \oplus \mathbf{C}_i)),
\end{equation}
where $\mathbf{W}_f \in \mathbb{R}^{d \times 4d}$ is a feature fusion matrix shared across all clip positions, and $\oplus$ denotes concatenation along the feature dimension. Finally, the fused video features $\tilde{\mathbf{V}}$ and text features $\mathbf{T}$ are provided to the encoder for cross-modal interaction. The output text token representations undergo max pooling to form a global summary $\mathbf{m}_g \in \mathbb{R}^d$, while the output video memory matrix $\mathbf{M} \in \mathbb{R}^{L_v \times d}$ serves as the input for subsequent temporal feature retrieval within the decoder.

\begin{figure}[t]
  \centering
  \includegraphics[width=\columnwidth]{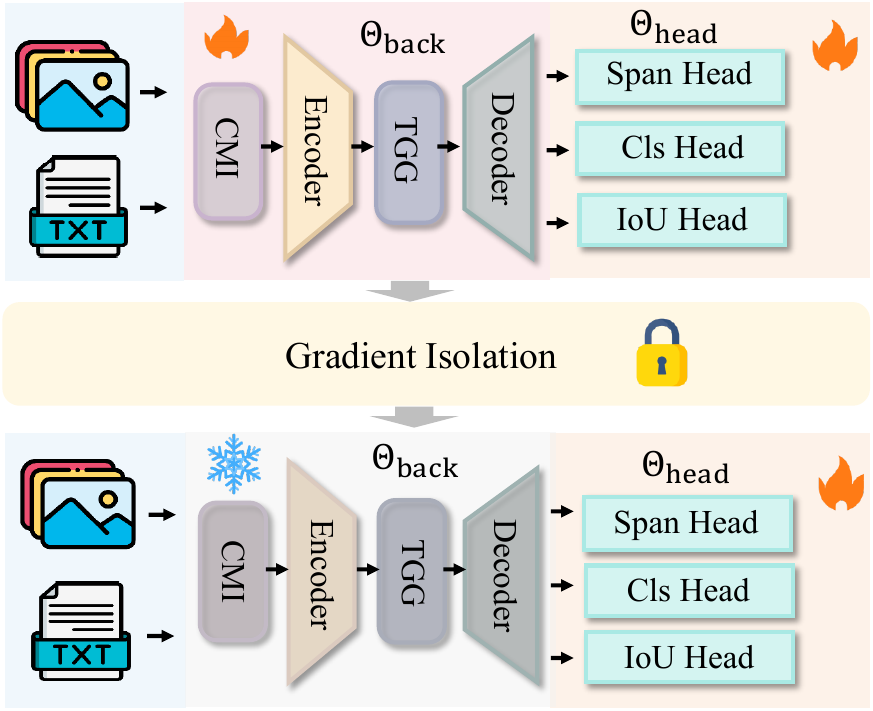}
  \caption{\textbf{ The gradient isolation strategy during post-training.} The parameter space is partitioned into a frozen backbone $\Theta_{\mathrm{back}}$ and an active detection head $\Theta_{\mathrm{head}}$ to ensure training stability. This configuration prevents feature manifold degradation when the model is subjected to high-variance policy gradients during the reinforcement learning phase.}
  \label{fig:post_training}
\end{figure}
\subsection{Semantic-Gated Query Decoding}
\textit{  1) Text-Guided Gating Mechanism.} A Text-Guided Gating mechanism is introduced to endow the task-agnostic decoder queries with the semantic prior of the query $\mathcal{Q}$. This operation occurs before the decoder, utilizing the global text summary $\mathbf{m}_g$ to inject semantic information into the initial query state $\mathbf{q}_\mathrm{init} \in \mathbb{R}^{N_q \times d}$. Specifically, cross-attention is performed between the initial query $\mathbf{q}_\mathrm{init}$ and the global text summary $\mathbf{m}_g$ to aggregate text context $\mathbf{q}_1 \in \mathbb{R}^{N_q \times d}$ for each query:
\begin{equation}
\mathbf{q}_1 = \mathrm{Attention}(\mathbf{q}_\mathrm{init}, \mathbf{m}_g, \mathbf{m}_g).
\end{equation}

To facilitate feature interaction among queries before injection, self-attention is applied to the residual sum of $\mathbf{q}_\mathrm{init}$ and $\mathbf{q}_1$. This provides each query vector with a rich representation $\mathbf{q}_2 \in \mathbb{R}^{N_q \times d}$ that integrates the global text prior. To adaptively control the injection amount of text information, data-dependent dynamic gating weights $\mathbf{g} \in (0,1)^{N_q \times d}$ are calculated based on the element-wise interaction between the query prior and the text context:
\begin{equation}
\mathbf{g} = \sigma(\mathbf{q}_\mathrm{init} \odot \mathbf{q}_1),
\end{equation}
where $\sigma(\cdot)$ is the Sigmoid function and $\odot$ denotes element-wise multiplication. The gating weight $\mathbf{g}$ reflects the alignment between the query prior and the text context across feature dimensions. The final query update is executed via residual gated injection and stabilized by layer normalization:
\begin{equation}
\tilde{\mathbf{q}}_\mathrm{init} = \mathrm{LN}(\mathrm{ReLU}(\mathbf{W}_g (\mathbf{g} \odot \mathbf{q}_2)) + \mathbf{q}_\mathrm{init}),
\end{equation}
where $\mathbf{W}_g \in \mathbb{R}^{d \times d}$ is a linear projection matrix. The gated and scaled $\mathbf{q}_2$ undergoes linear transformation and activation, and is then added to the original query prior $\mathbf{q}_\mathrm{init}$ as a residual connection. This design preserves the original learning capacity of the decoder while adaptively injecting text semantics.

\textit{  2) Temporal Feature Retrieval.} Subsequently, the decoder utilizes $\tilde{\mathbf{q}}_\mathrm{init}$ to retrieve temporal features from the memory matrix $\mathbf{M}$ via cross-attention. The decoder outputs a set of hidden states $\{\mathbf{h}_i\}_{i=1}^{N_q}$, which are directly forwarded to the detection head for the prediction of boundaries and categories.

\subsection{Supervised Warm-up and Gradient Isolation}
\textit{  1) Detection Head Structure.} Within the detection head, three parallel sub-heads process each hidden state $\mathbf{h}_i$ to generate the temporal span $\hat{\mathbf{s}}_i = (t_i, w_i)$, the classification logit $\hat{\mathbf{c}}_i$, and the localization quality score $\hat{u}_i$:
\begin{equation}
\hat{\mathbf{s}}_i = \Phi_\mathrm{span}(\mathbf{h}_i), \quad \hat{\mathbf{c}}_i = \Phi_\mathrm{cls}(\mathbf{h}_i), \quad \hat{u}_i = \sigma(\Phi_\mathrm{iou}(\mathbf{h}_i)),
\end{equation}
where $\Phi_\mathrm{span}$ and $\Phi_\mathrm{iou}$ denote three-layer multilayer perceptrons, $\Phi_\mathrm{cls}$ is a linear projection, and $\sigma(\cdot)$ is the Sigmoid function. For conciseness, the foreground logit is extracted from the two-dimensional classification logits $\hat{\mathbf{c}}_i$ and denoted as an independent foreground score $s_i$.

\textit{  2) Supervised Warm-up Objective.} During traditional supervised training, a bipartite matcher establishes an optimal one-to-one assignment between the $N_q$ predictions and the $K$ ground truth segments within the batch. Matched queries are assigned foreground targets, while the remainder are treated as background. Assuming $\mathcal{N}$ matching pairs are established with the assignment $\{(\sigma_i, \tau_i)\}_{i=1}^{\mathcal{N}}$, the supervision objective is formulated as:
\begin{equation}
\mathcal{L}_\mathrm{sup} = \lambda_\mathrm{L1}\mathcal{L}_1 + \lambda_\mathrm{giou}\mathcal{L}_\mathrm{giou} + \lambda_\mathrm{cls}\mathcal{L}_\mathrm{cls} + \lambda_\mathrm{iou}\mathcal{L}_\mathrm{iou}.
\end{equation}

The $\ell_1$ regression loss on the matched span parameters is defined as:
\begin{equation}
\mathcal{L}_1 = \frac{1}{\mathcal{N}}\sum_{i=1}^{\mathcal{N}} \|\hat{\mathbf{s}}_{\sigma_i} - \mathbf{s}_{\tau_i}\|_1.
\end{equation}

A scale-invariant structural penalty is imposed through the generalized tIoU loss:
\begin{equation}
\mathcal{L}_\mathrm{giou} = \frac{1}{\mathcal{N}}\sum_{i=1}^{\mathcal{N}} (1 - \mathrm{GIoU}(\hat{\mathbf{s}}_{\sigma_i}, \mathbf{s}_{\tau_i})).
\end{equation}

The quality head is supervised against the absolute error relative to the true Intersection-over-Union:
\begin{equation}
\mathcal{L}_\mathrm{iou} = \frac{1}{\mathcal{N}}\sum_{i=1}^{\mathcal{N}} |\hat{u}_{\sigma_i} - \mathrm{tIoU}(\hat{\mathbf{s}}_{\sigma_i}, \mathbf{s}_{\tau_i})|.
\end{equation}

The classification term $\mathcal{L}_\mathrm{cls}$ computes the weighted cross-entropy over all $N_q$ query slots, explicitly down-weighting the background class.

\textit{  3) Gradient Isolation Strategy.} As the training transitions to the post-training strategy, it is critical to recognize that while the aforementioned supervised objective $\mathcal{L}_\mathrm{sup}$ ensures initial convergence, the continuous proxy losses within it, namely $\mathcal{L}_\mathrm{giou}$ and $\mathcal{L}_1$, only provide stable gradients for global matching pairs. These losses cannot directly optimize the tIoU, which is a strictly non-differentiable metric highly dependent on the discrete ranking of foreground scores during evaluation. To bridge the misalignment between continuous proxy losses and discrete ranking metrics, and to prevent the disruption of the feature space caused by imposing high-variance policy gradients, the gradient computation for the backbone network is halted once $\mathcal{L}_\mathrm{sup}$ converges. As shown in Figure~\ref{fig:post_training}, the complete parameter space $\Theta$ is strictly partitioned into two disjoint sets:
\begin{equation}
\Theta_\mathrm{back} = \Theta_\mathrm{enc} \cup \Theta_\mathrm{proj} \cup \Theta_\mathrm{dec}, \quad \Theta_\mathrm{head} = \Theta_\mathrm{span} \cup \Theta_\mathrm{cls} \cup \Theta_\mathrm{iou}.
\end{equation}

The set $\Theta_\mathrm{back}$ is frozen, which restricts all subsequent optimization for the non-differentiable objective to the active set $\Theta_\mathrm{head}$. This operation ensures that the gradient flowing into the feature backbone is zero, establishing a stable state space for the progressive introduction of the policy.

\begin{figure}[t]
  \centering
  \includegraphics[width=\columnwidth]{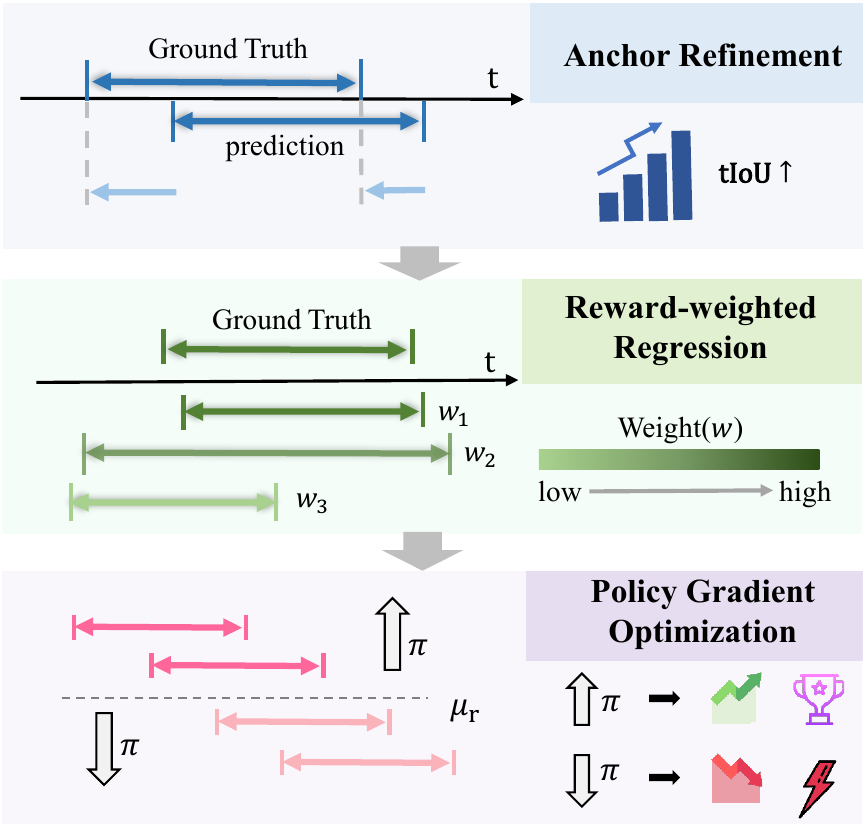}
  \caption{\textbf{ The TPRL paradigm.} This strategy aligns the training objective with the evaluation metric by transitioning from proxy losses to direct optimization. The process includes anchor refinement for calibration, reward-weighted regression for precision, and policy gradient optimization for final selection.}
  \label{fig:3phase}
\end{figure}

\subsection{Progressive Reinforcement Learning}
\textit{  1) Reinforcement Learning Paradigm.} With the feature fusion backbone frozen, the detection head undergoes refinement by gradually incorporating reinforcement learning. As depicted in Figure~\ref{fig:3phase}, this strategy involves three stages: anchor refinement, reward-weighted regression, and policy gradient optimization. All parameter updates are strictly confined within $\Theta_\mathrm{head}$.

\textit{  2) Anchor Refinement Stage.} The inference process retrieves the prediction with the highest rank based on the foreground score. This stage corrects the inconsistency between bipartite matching and evaluation selection through a supervised warm-up that specifically targets the highest-scoring prediction. Let $\kappa$ denote the query index satisfying:
\begin{equation}
s_\kappa = \max_{1 \le i \le N_q} s_i,
\end{equation}
and let $g^*$ be the ground truth segment that possesses the highest tIoU with $\hat{\mathbf{s}}_\kappa$. The corresponding loss function is defined as:
\begin{equation}
\mathcal{L}_\mathrm{ar} = \mathcal{L}_1(\hat{\mathbf{s}}_\kappa, g^*) + \mathcal{L}_\mathrm{giou}(\hat{\mathbf{s}}_\kappa, g^*).
\end{equation}

This function provides calibration for the top-scoring rank prior to the introduction of reward-based signals.

\textit{  3) Reward-Weighted Regression Stage.} This stage utilizes the actual tIoU reward $r_k$ as a weighting mechanism, which is applied exclusively to the top-$\hat{k}$ queries with the highest foreground scores to avoid gradient conflicts. The normalized weight $w_k$ is defined as:
\begin{equation}
w_k = \frac{r_k}{\sum_{j \in \mathcal{K}} r_j + \epsilon}.
\end{equation}

This mechanism guarantees that predictions yielding higher rewards receive larger gradients, further pushing aligned predictions toward precision. The differential allocation of gradients gradually widens the quality gap among the top-$\hat{k}$ queries. This process disperses the reward distribution for each query as it enters the third stage, thereby providing a stronger signal. The corresponding regression loss is calculated as:
\begin{equation}
\mathcal{L}_\mathrm{rwr} = \sum_{k \in \mathcal{K}} w_k \|\hat{\mathbf{s}}_k - g_k^*\|_1,
\end{equation}
where $g_k^*$ is the ground truth segment demonstrating the highest tIoU with the $k$-th prediction.

\begin{algorithm}[t]
\caption{Reinforcement Learning Formulation of PGO}
\label{alg:pgo_stage}
\begin{algorithmic}[1] 
\Statex \textbf{Input:} Video $V$, Text query $T$, Ground-truth set $\mathcal{G}$, Frozen backbone $\Theta_{\mathrm{back}}$, Active head $\Theta_{\mathrm{head}}$, Learning rate $\eta$
\Statex \textbf{Output:} Optimized prediction head parameters $\Theta_{\mathrm{head}}$

\State \textbf{Initialize} joint feature matrix  $\mathbf{M}$ and hidden states $\mathbf{h}$ via frozen backbone
\State \quad $\mathbf{M} \leftarrow \Phi_{\mathrm{back}}(V, T; \Theta_{\mathrm{back}})$
\State \quad $\mathbf{h}_i \leftarrow \Phi_{\mathrm{dec}}(\mathbf{M}, \tilde{\mathbf{q}}_{\mathrm{init}}; \Theta_{\mathrm{back}}), \quad \forall i \in \{1, \dots, N_q\}$

\For{each training epoch}
    \For{$i = 1$ to $N_q$}
        \State 
         $a_i = (\hat{\mathbf{s}}_i, \hat{\mathbf{c}}_i, \hat{u}_i) \leftarrow \Phi_{\mathrm{head}}(\mathbf{h}_i; \Theta_{\mathrm{head}})$
         \vspace{0.05in}
        \State
         $\pi(a_i \mid \mathbf{h}_i; \Theta_{\mathrm{head}}) \leftarrow \frac{\exp(\hat{\mathbf{c}}_i / \tau)}{\sum_{j=1}^{N_q} \exp(\hat{\mathbf{c}}_j / \tau)}$
         \vspace{0.05in}
        \State 
        $r_i \leftarrow \max_{g \in \mathcal{G}} \mathrm{tIoU}(\hat{\mathbf{s}}_i, g)$
    \EndFor

    \For{$i = 1$ to $N_q$}
        \State 
        $A_i \leftarrow \Phi_{\mathrm{norm}}\big(r_i \mid \{r_j\}_{j=1}^{N_q}\big)$
    \EndFor
    
    \State 
   $\mathcal{L}_{\mathrm{grpo}} \leftarrow \Phi_{\mathrm{loss}}\big(\{A_i\}_{i=1}^{N_q}, \pi\big)$
    \State 
    $\Theta_{\mathrm{head}} \leftarrow \Phi_{\mathrm{update}}\big(\Theta_{\mathrm{head}}, \mathcal{L}_{\mathrm{grpo}}, \eta\big)$
\EndFor
\end{algorithmic}
\end{algorithm}

\textit{  4) Policy Gradient Optimization Stage.} To frame Policy Gradient Optimization (PGO) as a reinforcement learning procedure detailed in Algorithm \ref{alg:pgo_stage}, we model boundary prediction as a Markov Decision Process (MDP). $\mathcal{E}$ explicitly freezes the backbone $\Theta_{\mathrm{back}}$ to prevent high-variance gradients from corrupting multimodal representations. Accordingly, the stable state space $\mathcal{S}$ is composed of the decoder's output hidden states $\mathbf{h}_i$. The detection head $\Theta_{\mathrm{head}}$ acts as the agent, processing each state $\mathbf{h}_i$ to generate an action $a_i \in \mathcal{A}$ under policy $\pi$. Here, $a_i$ comprises the candidate temporal span, class logits, and quality score.

Instead of proxy losses, $\mathcal{E}$ directly optimizes the non-differentiable tIoU against the ground-truth set $\mathcal{G}$, which is strictly defined as the reward $\mathcal{R}$. Adopting Group Relative Policy Optimization (GRPO) \cite{shao2024deepseekmath}, we compute the relative advantage $A_i$ within the $N_q$ queries to maximize this metric:

\begin{equation}
A_i = \frac{r_i - \mu_r}{\sigma_r + \epsilon},
\end{equation}
where $r_i$ denotes the individual reward, $\mu_r$ is the intra-group reward mean, $\sigma_r$ stands for the standard deviation, and $\epsilon$ is a small constant. Because background queries inherently yield a zero reward, the gradient naturally concentrates on the minority of high-reward foreground queries, thereby driving the detection head to output the optimal prediction. The policy loss $\mathcal{L}_\mathrm{grpo}$ and the final objective $\mathcal{L}_\mathrm{pgo}$ are thus formulated as:

\begin{equation}
\mathcal{L}_\mathrm{grpo} = -\sum_{i=1}^{N_q} A_i\log \pi(a_i \mid \mathbf{h}_i; \Theta_{\mathrm{head}}),
\end{equation}

\begin{equation}
\mathcal{L}_\mathrm{pgo} = \mathcal{L}_\mathrm{grpo} + \lambda_\mathrm{rw}\mathcal{L}_\mathrm{rwr}.
\end{equation}

By progressively introducing RL across three stages, the optimization objective of the detection head shifts from proxy losses to the direct optimization of the tIoU, perfectly aligning the training objective with the evaluation metric.

\section{Experiments}

%%%%%%%%%%%%%%%%%%%%%%%%%%%%%%%%%%%%%%%%%%%%这里没问题
\subsection{Datasets}

Following the experimental settings of existing temporal video grounding methods, we conduct experiments on three datasets: Charades-STA~\cite{gao2017tall}, QVHighlights~\cite{lei2021detecting}, and TACoS~\cite{regneri2013grounding}, to verify the effectiveness of the proposed method across scenarios and localization difficulties. All experiments are conducted on a single NVIDIA RTX 4090 GPU.

\textit{  1) Charades-STA:} Built upon Charades, this temporal sentence localization dataset primarily comprises indoor daily activity videos. The dataset contains 6,672 videos, which are separated into training and test sets of 5,338 and 1,334 videos, respectively, with an average video length of about 29.8 seconds. It consists of 16,128 video-query pairs, including 12,408 for training and 3,720 for testing. Additionally, each multi-text query corresponds to one video and contains an average of 2.41 sentences. This dataset is commonly used to evaluate a model's temporal localization capability in daily behavior scenarios.

\textit{  2) QVHighlights:} This is a popular dataset for both video moment retrieval and highlight detection based on natural language queries. Specifically, it consists of 18,367 moments in 10,148 open-domain videos, each of which lasts approximately 150 seconds and is paired with a human-written text query. The dataset provides not only relevant moment annotations but also highlight saliency scores, thereby comprehensively evaluating the model's capabilities in video segment localization, candidate ranking, and highlight recognition. The dataset is divided into training, validation, and test sets. To ensure fairness, evaluation on its test set can only be performed by submitting predictions to its online server.

\textit{  3) TACoS:} This is a fine-grained temporal localization dataset for cooking scenarios, which combines the MPII corpus with kitchen scene videos, resulting in 127 videos focused on cooking activities. The average video length in this dataset is relatively long, at about 4.79 minutes, and each multi-text query consists of an average of 8.75 sentences. The training, validation, and test sets contain 1,107, 418, and 380 video-multi-text query pairs, respectively. Because different actions typically occur in similar scenes and their temporal boundaries are subtle, this dataset imposes higher requirements on the model's fine-grained semantic understanding and temporal boundary localization capabilities.

\begin{table}[t]
\scriptsize
\caption{Comparison of the Proposed Method with State-of-the-Art Methods on the Charades-STA Dataset. Evaluation Results for R1@0.3, R1@0.5, R1@0.7, and mIoU Are Listed. The Best and Second-Best Values Are Highlighted in \textbf{Bold} and \underline{Underline}.}
\label{tab:comparison_charades_sta}
\centering
\setlength{\tabcolsep}{2pt}
\resizebox{\columnwidth}{!}{
\begin{tabular}{lcccc}
\toprule
Method & R1@0.3 & R1@0.5 & R1@0.7 & mIoU  \\
\midrule
Moment-DETR (NeurIPS21)~\cite{lei2021detecting}             & 65.83 & 52.07 & 30.59 & 45.54 \\
STCNet (WWW23)~\cite{wang2023spatiotemporal}                & -     & 59.09 & \underline{38.68} & 52.18 \\
MomentDiff (NeurIPS23)~\cite{li2023momentdiff}              & -     & 55.57 & 32.42 & -     \\
QD-DETR (CVPR23)~\cite{moon2023query}                       & -     & 57.31 & 32.55 & -     \\
UniVTG (ICCV23)~\cite{lin2023univtg}                        & 70.81 & 58.01 & 35.65 & 50.10 \\
TR-DETR (AAAI24)~\cite{sun2024tr}                           & -     & 57.61 & 33.52 & -     \\
LLMEPET (ACM MM24)~\cite{jiang2024prior}                    & 70.91 & -     & 36.49 & 50.25 \\
UVCOM (CVPR24)~\cite{xiao2024bridging}                      & -     & 59.25 & 36.64 & -     \\
BAM-DETR (ECCV24)~\cite{lee2024bam}                         & \underline{72.93} & \underline{59.95} & \textbf{39.38} & \underline{52.33} \\
CG-DETR (PR25)~\cite{Moon2025CorrelationguidedCO}           & 70.40 & 58.40 & 36.30 & 50.10 \\
RGTR (AAAI25)~\cite{sun2025diversifying}                    & 72.04 & 57.93 & 35.16 & 50.32 \\
\midrule
\rowcolor{baselinerow}
GIRL-DETR (Ours)                                                & \textbf{75.54} & \textbf{61.75} & 37.47 & \textbf{52.82} \\
\bottomrule
\end{tabular}
}
\end{table}

\begin{table}[t]
\scriptsize
\caption{Comparison of the Proposed Method with State-of-the-Art Methods on the QVHighlights Dataset. Results for R1@0.5, R1@0.7, mAP@0.5, and mAP Are Listed. The Best and Second-Best Values Are Highlighted in \textbf{Bold} and \underline{Underline}.}
\label{tab:comparison_qvhighlights}
\centering
\setlength{\tabcolsep}{2pt}
\resizebox{\columnwidth}{!}{
\begin{tabular}{lcccc}
\toprule
Method & R1@0.5 & R1@0.7 & mAP@0.5 & mAP \\
\midrule
Moment-DETR (NeurIPS21)~\cite{lei2021detecting}        & 52.89 & 33.02 & 54.82 & 30.73 \\
MomentDiff (NeurIPS23)~\cite{li2023momentdiff}         & 58.21 & 41.48 & 54.57 & 36.84 \\
UniVTG (ICCV23)~\cite{lin2023univtg}                   & 58.86 & 40.86 & 57.60 & 35.47 \\
QD-DETR (CVPR23)~\cite{moon2023query}                  & 62.40 & 44.98 & 63.17 & 41.44 \\
MRNet (ACM MM24)~\cite{hu2024maskable}                 & 64.85 & 46.63 & 65.11 & 41.63 \\
TR-DETR (AAAI24)~\cite{sun2024tr}                      & 64.66 & 48.96 & 63.98 & 42.62 \\
BAM-DETR (ECCV24)~\cite{lee2024bam}                    & 62.71 & 48.64 & 64.57 & 45.36 \\
CG-DETR (PR25)~\cite{Moon2025CorrelationguidedCO}      & 65.40 & 48.40 & 64.50 & 42.90 \\
RGTR (AAAI25)~\cite{sun2025diversifying}               & \underline{65.50} & \underline{49.22} & \underline{67.12} & 45.53 \\
SA-DETR (COLING25)~\cite{xiong2025sa}                  & 64.96 & 49.09 & 65.30 & \underline{47.40} \\
\midrule
\rowcolor{baselinerow}
GIRL-DETR (Ours)                                       & \textbf{73.03} & \textbf{56.77} & \textbf{69.85} & \textbf{50.10} \\
\bottomrule
\end{tabular}
}
\end{table}

\begin{table}[t]
\scriptsize
\caption{Comparison of the Proposed Method with State-of-the-Art Methods on the TACoS Dataset. Evaluation Results for R1@0.3, R1@0.5, R1@0.7, and mIoU Are Listed. The Best and Second-Best Values Are Highlighted in \textbf{Bold} and \underline{Underline}.}
\label{tab:comparison_tacos}
\centering
\setlength{\tabcolsep}{2pt}
\resizebox{\columnwidth}{!}{
\begin{tabular}{lcccc}
\toprule
Method & R1@0.3 & R1@0.5 & R1@0.7 & mIoU  \\
\midrule
QAVE (Neurocomputing22)~\cite{hao2022query}            & 50.11 & 38.57 & \underline{27.52} & 37.45 \\
STCNet (WWW23)~\cite{wang2023spatiotemporal}           & 38.84 & 25.42 & -     & 26.25 \\
MomentDiff (NeurIPS23)~\cite{li2023momentdiff}         & 44.78 & 33.68 & -     & -     \\
UniVTG (ICCV23)~\cite{lin2023univtg}                   & 51.44 & 34.97 & 17.35 & 33.60 \\
LLMEPET (ACM MM24)~\cite{jiang2024prior}               & 52.73 & -     & 22.78 & 36.55 \\
UVCOM (CVPR24)~\cite{xiao2024bridging}                 & -     & 36.39 & 23.32 & -     \\
BAM-DETR (ECCV24)~\cite{lee2024bam}                    & 56.69 & 41.54 & 26.77 & 39.31 \\
MRNet (ACM MM24)~\cite{hu2024maskable}                 & 56.16 & 41.31 & -     & 39.45 \\
CG-DETR (PR25)~\cite{Moon2025CorrelationguidedCO}      & 54.40 & 39.50 & 23.40 & 37.40 \\
RGTR (AAAI25)~\cite{sun2025diversifying}               & 53.04 & 40.31 & 24.32 & 37.44 \\
SA-DETR (COLING25)~\cite{xiong2025sa}                  & \textbf{58.16} & \underline{42.56} & \textbf{27.87} & \underline{40.03} \\
\midrule
\rowcolor{baselinerow}
GIRL-DETR (Ours)                                           & \underline{56.71} & \textbf{43.69} & 26.69 & \textbf{40.20} \\
\bottomrule
\end{tabular}
}
\end{table}
\subsection{ Implementation Details}

The following experimental settings are primarily based on the Charades-STA dataset. Minor dataset-specific parameter variations can be found in our open-source code.

\textit{  1) Supervised Warm-up.} We utilize the ViT-B~\cite{dosovitskiy2020image} architecture from CLIP~\cite{radford2021learning}, the base model of BLIP~\cite{li2022blip}, and the isolated ViT-H visual backbone of InternVideo2~\cite{wang2024internvideo2} to extract video features. To establish a rigorous and relatively fair comparison, we also evaluate the conventional SlowFast~\cite{feichtenhofer2019slowfast} network. By strictly limiting InternVideo2 to its vision encoder rather than deploying the full multimodal suite, we maintain a parameter scale commensurate with heavy visual backbones, ensuring that performance evaluations focus purely on the quality of spatial-temporal representations. Maximum video length is 75. Transformer encoder and decoder both consist of 3 layers with a 256 hidden dimension. We employ AdamW optimizer with a batch size of 32 and a 2e-4 base learning rate, decaying at the 100th epoch. Contrastive alignment and hard positive negative loss coefficients are 0.3 and 10, with an EMA decay of 0.999.

\textit{  2) RL Post-training.} After supervised convergence, we freeze the backbone and fine-tune the detection head with a 32 batch size and a 5e-5 learning rate. We implement a Three-stage Progressive Reinforcement Learning (TPRL) strategy including Anchor Refinement (AR) stage, Reward-Weighted Regression (RWR) stage, and Policy Gradient Optimization (PGO) stage. Training lasts 30 epochs, with 10 epochs per stage. Parameters include a 0.03 base RL coefficient, a 1.0 top-1 refinement coefficient, and a cosine decay schedule for policy gradients.

\textit{  3) Inference and Testing.} During inference, the final confidence score is modulated by the product of foreground probability and predicted IoU. The model directly outputs ranked temporal span candidates based on this score. To measure retrieval and localization precision, we adopt Recall at various thresholds of the tIoU, specifically R1@0.3, R1@0.5, and R1@0.7. Additionally, we report the mean temporal IoU (mIoU), mAP@0.5, and mAP@Avg, hereafter denoted simply as mAP.

%%%%%%%%%%%%%%%%%%%%%%%%%%%%%%%%%%%%%%%%%%%%

\subsection{Comparison with State-of-the-Art Methods}

To evaluate the effectiveness of the proposed GIRL-DETR, we compare it with state-of-the-art temporal grounding methods on three public benchmarks, namely QVHighlights, Charades-STA, and TACoS. The results show that GIRL-DETR achieves competitive performance across different datasets and evaluation metrics, which indicates that the Reinforcement Learning (RL) mechanism based on tIoU rewards effectively reduces the mismatch between conventional proxy regression losses and final evaluation metrics.

\textit{  1) Comparison on Charades-STA:} GIRL-DETR achieves the best results on R1@0.3 and R1@0.5, reaching 75.54\% and 61.75\%, respectively, and achieves a remarkable mIoU of 52.82\%. Compared with UniVTG~\cite{lin2023univtg}, RGTR~\cite{sun2025diversifying}, and BAM-DETR~\cite{lee2024bam}, our method performs better in medium- and low-threshold recall as well as overall overlap quality. Compared with the strongest baseline in our comparison, it improves R1@0.3 and R1@0.5 by 2.61\% and 1.80\%, respectively, and outperforms the baseline by 0.49\% on mIoU, as shown in Table~\ref{tab:comparison_charades_sta}. Although the R1@0.7 of GIRL-DETR is slightly lower than that of BAM-DETR, obtaining modest improvements over baselines on mIoU and mainstream recall thresholds shows that the model localizes semantically relevant segments more stably, rather than obtaining a local advantage only under a single strict threshold. 

\textit{  2) Comparison on QVHighlights:} GIRL-DETR achieves the best results on all metrics, with R1@0.5, R1@0.7, mAP@0.5, and mAP reaching 73.03\%, 56.77\%, 69.85\%, and 50.10\%, respectively. Compared with strong prior methods such as RGTR, SA-DETR~\cite{xiong2025sa}, and BAM-DETR, our method shows clear advantages in both recall and mAP. Compared with the strongest baseline in our comparison, R1@0.5 and R1@0.7 improve by 7.53\% and 7.55\%, respectively, as shown in Table~\ref{tab:comparison_qvhighlights}. R1@0.7 is a stricter localization metric. Its improvement indicates that GIRL-DETR not only recalls target moments more accurately, but also maintains more reliable boundary prediction and confidence ranking under high-overlap requirements.

\textit{  3) Comparison on TACoS:} GIRL-DETR achieves the best results on R1@0.5, reaching 43.69\%, and yields a mIoU of 40.20\%. Compared with recent methods such as SA-DETR, BAM-DETR, RGTR, and CG-DETR~\cite{Moon2025CorrelationguidedCO}, our method leads on the representative medium-threshold metric and the average overlap quality. Compared with the strongest baseline in our comparison, it improves R1@0.5 by 1.13\% and outperforms the baseline by 0.17\% on mIoU, as shown in Table~\ref{tab:comparison_tacos}. Since TACoS contains many cooking activities with visually similar scenes and fine-grained action boundaries, models are easily affected by semantic confusion between adjacent actions and ambiguous boundaries. Therefore, a higher R1@0.5 and a better mIoU on this dataset show that GIRL-DETR maintains localization robustness in complex fine-grained scenarios.

\definecolor{baselinerow}{gray}{0.9} % 如果没定义背景色，请取消此行注释

\begin{table}[t]
\centering
\caption{Effectiveness of the Proposed Modules on the Validation Split of QVHighlights.}
\label{tab:ablation}
\setlength{\tabcolsep}{2pt}
\resizebox{\columnwidth}{!}{%
\begin{tabular}{ccccccccccc}
\toprule % 标准学术顶线（粗线）
No. & DETR & CMI & TGG & TPRL & SF & IV2 & R1@0.5 & R1@0.7 & mIoU & mAP \\ 
\midrule % 标准学术栏目线（细线）
1 & \checkmark & $\times$ & $\times$ & $\times$ & $\times$ & \checkmark 
  & 71.35 & 55.16 & 64.59 & 47.28 \\

2 & \checkmark & \checkmark & $\times$ & $\times$ & $\times$ & \checkmark 
  & 71.68 & 56.39 & 65.67 & 48.89 \\

3 & \checkmark & \checkmark & \checkmark & $\times$ & $\times$ & \checkmark 
  & 72.13 & 56.77 & 66.03 & 49.38 \\

4 & \checkmark & \checkmark & \checkmark & \checkmark & \checkmark & $\times$ 
  & 71.23 & 55.94 & 65.81 & 47.82 \\

5 & \checkmark & \checkmark & \checkmark & \checkmark & $\times$ & \checkmark 
  & \textbf{73.03} & \textbf{56.77} & \textbf{66.53} & \textbf{50.10} \\
\bottomrule % 标准学术底线（粗线）
\end{tabular}%
}
\vspace{2pt} 
\flushleft \footnotesize % 使用更小的字号，并靠左对齐
\textit{Note:} SF and IV2 denote features extracted by SlowFast and InternVideo2.
\label{tab:ablation}
\end{table}

\subsection{Ablation on Proposed Modules}

To verify the contribution of our proposed core modules, we conduct progressive module ablation experiments on QVHighlights. The results are shown in Table~\ref{tab:ablation}. The baseline model, which utilizes the basic DETR framework, yields R1@0.5, R1@0.7, and mAP of 71.35\%, 55.16\%, and 47.28\%, respectively, and achieves a mIoU of 64.59\%. Introducing Cross-Modal Interaction (CMI) brings stable improvements across all metrics. For instance, R1@0.7 increases from 55.16\% to 56.39\%, and mAP increases from 47.28\% to 48.89\%, demonstrating that explicitly modeling fine-grained interactions helps reduce unimodal limitations. Building upon this, integrating our TGG module further boosts R1@0.5 and mAP to 72.13\% and 49.38\%, respectively, and brings a performance improvement with a mIoU of 66.03\%. This validates our motivation that Text-Guided Gating effectively injects textual concepts into decoder queries based on query semantics, successfully suppressing redundant segment responses.

Following the cross-modal modeling with CMI and TGG, we evaluate another core contribution, the TPRL optimization strategy. Applying this strategy to the representations derived from the InternVideo2 ViT-H backbone achieves the optimal performance, with R1@0.5, R1@0.7, and mAP reaching 73.03\%, 56.77\%, and 50.10\%, respectively, and yields a remarkable mIoU of 66.53\%. To rigorously demonstrate that these performance gains stem from our proposed architecture rather than merely relying on a heavy feature extractor, we conduct an experiment where the input video representations are replaced with those from the conventional SlowFast network widely used in previous works. Even when utilizing the SlowFast features, the complete framework maintains competitive performance, achieving R1@0.5 and mAP of 71.23\% and 47.82\%, respectively, and obtains a mIoU of 65.81\%. This robust performance across different feature backbones confirms the intrinsic, feature-agnostic effectiveness of our proposed modules. It proves that while the InternVideo2 representations provide a highly discriminative semantic basis, our core designs are fundamentally responsible for optimizing boundary prediction and ranking quality.

\begin{figure}[t]
  \centering
  \includegraphics[width=\columnwidth]{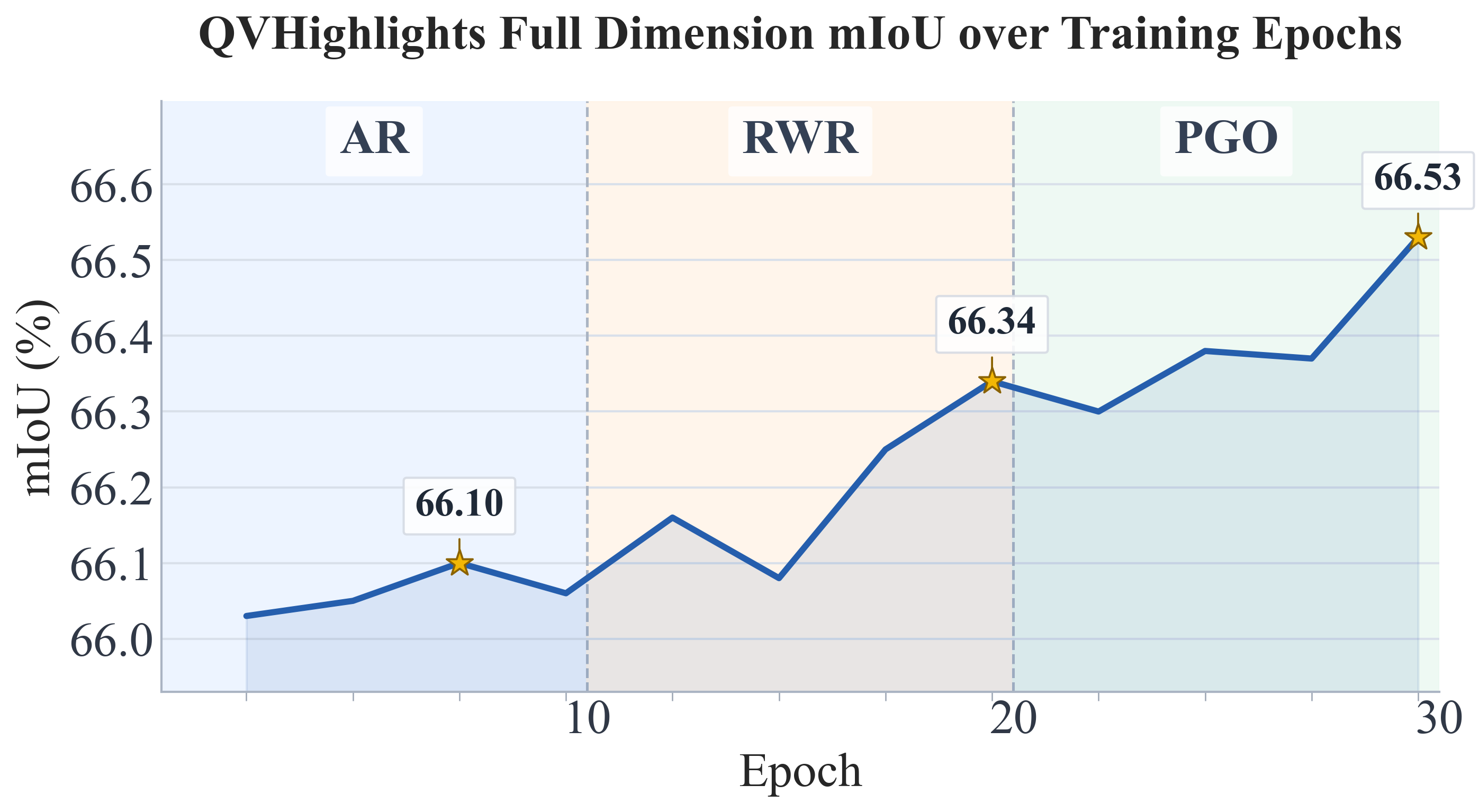}
  \caption{\textbf{Performance evolution of TPRL on QVHighlights.} The steady improvement of the core metric mIoU across the three stages demonstrates the effectiveness of TPRL in enhancing temporal localization.}
  \label{fig:3phase_rise}
\end{figure}

\subsection{Rationality of the TPRL Design}

To further analyze the training dynamics of the TPRL strategy, we report the variation of mIoU over training epochs on QVHighlights, as shown in Fig.~\ref{fig:3phase_rise}. The whole post-training process is divided into three stages: AR, RWR, and PGO. It can be observed that at the conclusion of the initial AR stage, the model establishes a solid performance baseline, indicating that it can already achieve measurable improvements over supervised training after preliminary action adaptation and prediction head calibration. After entering the second stage, the curve exhibits a stair-step progression, demonstrating that reward-weighted optimization can progressively emphasize predictions with higher tIoU and guide the model from conventional proxy losses toward an optimization objective that is more consistent with the final evaluation metric. In the third PGO stage, the model achieves its peak performance across the entire training process by directly maximizing the non-differentiable localization metric through relative strategy optimization. Notably, although the curve exhibits minor local fluctuations inherent to RL policy exploration, it maintains an overall upward trend rather than severe oscillations. This suggests that our method does not simply apply RL directly to the temporal localization model. Instead, it progressively releases the effect of the reward signal through stage-wise constraints. This process effectively reduces the risk of high-variance gradients in the early stage of RL training, allowing the model to first stabilize the prediction distribution and then further improve boundary quality with the tIoU reward. Therefore, this experiment verifies the necessity of the three-stage design from the perspective of training dynamics: AR stabilizes the prediction space, RWR introduces reward preference, and PGO further strengthens policy optimization aligned with the final localization metric.

\definecolor{darkgreen}{RGB}{0, 128, 0} 
\begin{table*}[t]
\centering
\setlength{\tabcolsep}{9pt}
\caption{Generalization of the Proposed TPRL Strategy across Various VMR Baselines on QVHighlights.}
\label{tab:three_stage_rl_qvhighlights}
\resizebox{\textwidth}{!}{
\begin{tabular}{llccccc}
\toprule
\textbf{Method} & \textbf{Module} & \textbf{R1@0.3} & \textbf{R1@0.5} & \textbf{R1@0.7} & \textbf{mIoU} & \textbf{mAP} \\
\midrule
\multirow{2}{*}{VideoLights~\cite{paul2024videolights}} 
& Baseline & 80.06 & 70.00 & 55.29 & 64.26 & 48.44 \\
& w/ TPRL     & 81.03 \textcolor{darkgreen}{(+0.97)} & 70.39 \textcolor{darkgreen}{(+0.39)} & 56.39 \textcolor{darkgreen}{(+1.10)} & 64.96 \textcolor{darkgreen}{(+0.70)} & 49.40 \textcolor{darkgreen}{(+0.96)} \\
\midrule
\multirow{2}{*}{UniVTG~\cite{lin2023univtg}} 
& Baseline & 71.81 & 59.74 & 40.90 & 53.36 & 32.59 \\
& w/ TPRL     & 73.23 \textcolor{darkgreen}{(+1.42)} & 60.26 \textcolor{darkgreen}{(+0.52)} & 41.55 \textcolor{darkgreen}{(+0.65)} & 54.18 \textcolor{darkgreen}{(+0.82)} & 34.20 \textcolor{darkgreen}{(+1.61)} \\
\midrule
\multirow{2}{*}{QD-DETR~\cite{moon2023query}} 
& Baseline & 74.77 & 62.90 & 46.77 & 57.59 & 41.24 \\
& w/ TPRL     & 75.74 \textcolor{darkgreen}{(+0.97)} & 63.74 \textcolor{darkgreen}{(+0.84)} & 48.13 \textcolor{darkgreen}{(+1.36)} & 58.34 \textcolor{darkgreen}{(+0.75)} & 42.36 \textcolor{darkgreen}{(+1.12)} \\
\midrule
\multirow{2}{*}{CG-DETR~\cite{Moon2025CorrelationguidedCO}} 
& Baseline & 77.94 & 67.35 & 52.06 & 61.69 & 44.93 \\
& w/ TPRL     & 78.52 \textcolor{darkgreen}{(+0.58)} & 67.61 \textcolor{darkgreen}{(+0.26)} & 52.45 \textcolor{darkgreen}{(+0.39)} & 62.23 \textcolor{darkgreen}{(+0.54)} & 45.49 \textcolor{darkgreen}{(+0.56)} \\
\bottomrule
\end{tabular}
}
\end{table*}

\subsection{Generalization of TPRL Across Various VMR Models}

Table~\ref{tab:three_stage_rl_qvhighlights} presents the performance of various temporal video grounding models on the QVHighlights dataset before and after the application of our three-stage TPRL strategy. In these experiments, we freeze the pre-trained backbones of the baseline models and only train the prediction heads using TPRL as a post-training strategy to directly optimize the localization results. As shown in the table, the introduction of TPRL yields consistent improvements across all evaluated metrics, including R1 at different overlap thresholds, mIoU, and mAP, for all four baseline methods.

UniVTG achieves the highest improvements on several metrics, with R1@0.3 and mAP increasing by 1.42\% and 1.61\% respectively. QD-DETR~\cite{moon2023query} also obtains remarkable enhancements, particularly a 1.36\% increase on the strict R1@0.7 and a 0.84\% boost on R1@0.5. Additionally, VideoLights~\cite{paul2024videolights} shows a 1.10\% improvement on R1@0.7, while CG-DETR observes steady enhancements across all metrics.

These stable performance gains, achieved by exclusively updating the prediction heads, validate that proxy loss degradation is a common issue in many VMR models. Furthermore, the quantitative results explicitly demonstrate that our TPRL serves as an effective post-training strategy to alleviate this degradation and improve temporal grounding performance.

\subsection{Qualitative Analysis of the TPRL Strategy}

\begin{figure*}[t]
 \centering
\includegraphics[width=\linewidth]{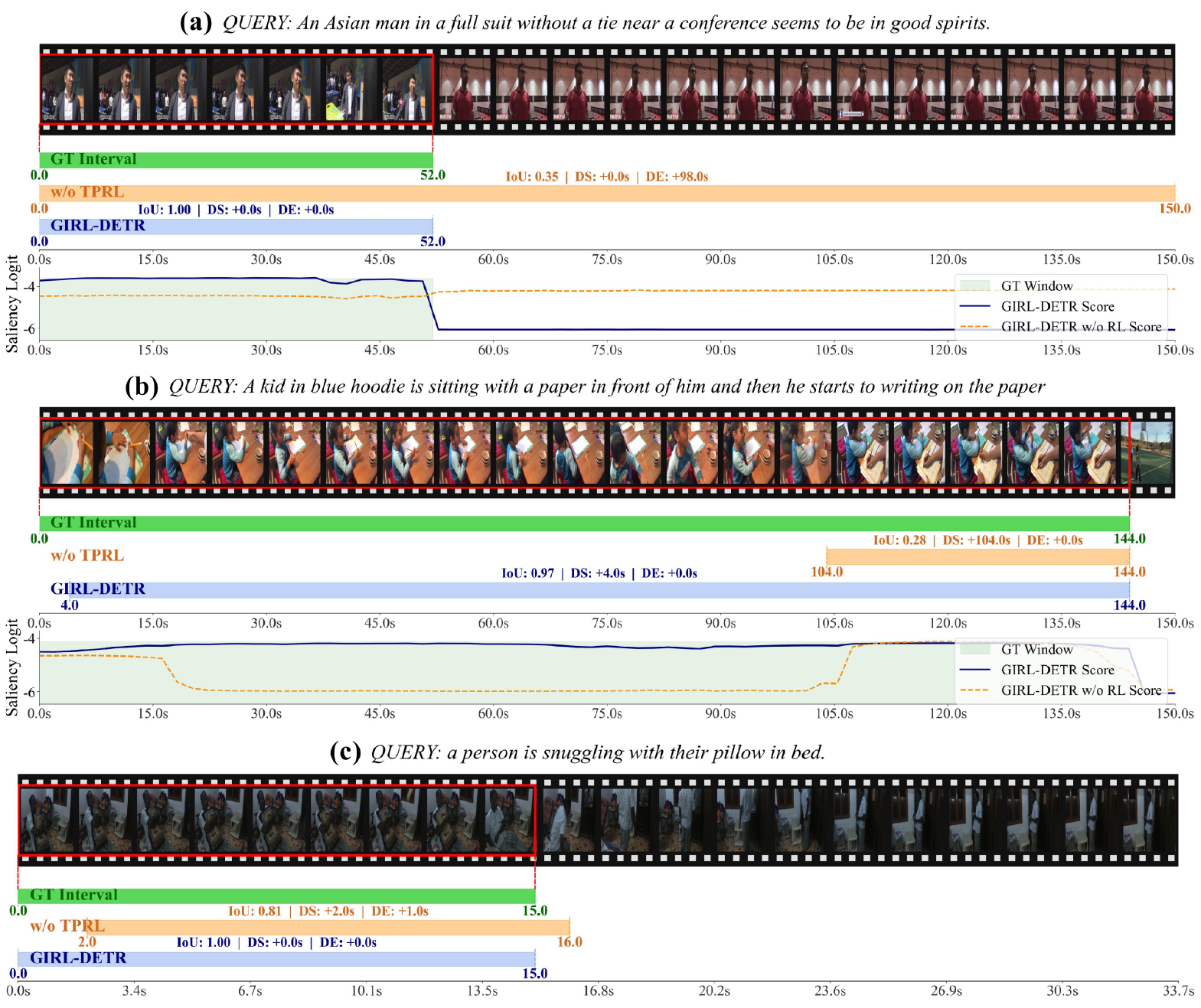}
 \caption{\textbf{Qualitative comparison of localization results on QVHighlights and Charades-STA.} GIRL-DETR effectively rectifies the severe boundary misalignment, observed in the baseline without TPRL, achieving highly accurate boundary regression and improving the confidence score distribution.}
\label{fig:plot}
\end{figure*}

To demonstrate the effectiveness of the proposed GIRL-DETR, we visualize qualitative localization results in Fig.~\ref{fig:plot}. Specifically, the first two cases represent queries from the QVHighlights dataset. The baseline model, trained solely with proxy losses and without TPRL, frequently exhibits severe boundary misalignment, such as boundary over-extension or premature truncation. By incorporating TPRL, GIRL-DETR calibrates these temporal shifts and achieves precise boundary regression. Furthermore, the corresponding confidence score curves indicate that the TPRL strategy corrects the ranking degradation caused by the loss-metric mismatch, ensuring that the model assigns higher confidence strictly to accurately localized moments. To verify the generalization capability of the proposed method across different video domains, the third case provides an additional retrieval example from the Charades-STA dataset. The consistent performance improvements across these distinct benchmarks demonstrate the robustness of the framework in handling various video contexts, thereby establishing GIRL-DETR as a reliable solution for fine-grained temporal localization.

\begin{table}[t]
\centering
\caption{Stage-Wise Ablation Study on the Optimization Stability of the TPRL Strategy across the Charades-STA and QVHighlights Datasets.}
\label{tab:three_stage_ablation}
\setlength{\tabcolsep}{2pt}
\resizebox{\columnwidth}{!}{%
\begin{tabular}{lccccccc} 
\toprule % 标准学术顶线（粗线）
Datasets & AR & RWR & PGO & R1@0.5 & R1@0.7 & mIoU & mAP \\
\midrule % 标准学术栏目线（细线）
\multirow{3}{*}{Charades-STA} 
& $\times$ & $\times$ & \checkmark 
& 60.81 & 36.61 & 52.24 & 33.96 \\
& $\times$ & \checkmark & \checkmark 
& 61.02 & 36.85 & 52.43 & 34.51 \\

& \checkmark & \checkmark & \checkmark 
& \textbf{61.75} & \textbf{37.47} & \textbf{52.82} & \textbf{36.32} \\
\midrule % 分隔不同数据集的细线
\multirow{3}{*}{QVHighlights} 
& $\times$ & $\times$ & \checkmark 
& 71.16 & 55.48 & 65.26 & 45.93 \\
& $\times$ & \checkmark & \checkmark 
& 72.26 & 55.74 & 65.50 & 45.80 \\

& \checkmark & \checkmark & \checkmark 
& \textbf{73.03} & \textbf{56.77} & \textbf{66.53} & \textbf{50.10} \\
\bottomrule % 标准学术底线（粗线）
\end{tabular}%
}
\end{table}

%%%%%%%%%%%%%%%%%%%%%%%%%%%%%%%%%%%%%%%%%%%%%%%%%%%%%%

\subsection{Ablation on TPRL Optimization Stability}

To verify the necessity of the progressive transition from supervised training to RL, we evaluate the optimization stability and performance of TPRL across various configurations on Charades-STA and QVHighlights. Notably, to ensure a fair comparison, all reported scores represent the best performance in the PGO stage with the same number of epochs across experiments. As shown in Table~\ref{tab:three_stage_ablation}, the complete TPRL setting, which integrates the reinforcement learning reward and penalty components AR, RWR, and PGO, achieves an R1@0.5 of 61.75\% and a mIoU of 52.82\% on Charades-STA, and an R1@0.5 of 73.03\% and a mIoU of 66.53\% on QVHighlights.

When the initial AR stage is removed, leaving only the RWR and PGO stages, a clear performance degradation is observed. Specifically, on Charades-STA, R1@0.5 and mIoU drop to 61.02\% and 52.43\%, respectively. Similarly, on QVHighlights, R1@0.5 decreases to 72.26\% and mAP drops to 45.80\%. This decline demonstrates that without the AR stage to provide action-space adaptation and a smooth transition from supervised manifolds, the model lacks a stable initialization for subsequent reward-driven optimization. In the absence of this anchor calibration, the abrupt shift from traditional proxy losses to RL signals causes the convergence to fall into sub-optimal solutions, confirming that AR is essential for stabilizing the prediction distribution before introducing reward-based signals.

The necessity of the intermediate RWR stage is further highlighted when comparing the PGO only configuration with the RWR and PGO variant as well as the full TPRL setting. Removing RWR and AR simultaneously causes the performance to reach its lowest point; on Charades-STA, R1@0.5 and mIoU fall to 60.81\% and 52.24\%, while on QVHighlights, R1@0.5 and mIoU drop to 71.16\% and 65.26\%. This further degradation reveals that directly applying high-variance policy gradients to the lightweight prediction head without the progressive guidance of AR and RWR leads to severe gradient oscillation or even non-convergence. The progressive design of TPRL effectively mitigates these risks: AR stabilizes the prediction space, RWR establishes a robust reward preference, and PGO finally executes a complete reinforcement learning reward and penalty mechanism based on relative advantages within a stabilized policy space to further improve the localization performance. These results strongly validate that each stage of TPRL is essential to ensure training stability and to effectively introduce RL into lightweight VMR models.

%%%就差这一节了

\begin{table}[t]
\centering
\caption{Ablation Study of the Gradient-Isolation Strategy on the QVHighlights Dataset.}
\label{tab:unfrozen_ablation}
\setlength{\tabcolsep}{3.5pt}
\resizebox{\columnwidth}{!}{%
\begin{tabular}{cccccccccc} % 移除了竖线，共计 10 列
\toprule % 标准学术顶线（粗线）
No. & Head & TGG & CMI & FLED & Others & R1@0.5 & R1@0.7 & mIoU & mAP \\
\midrule % 标准学术栏目线（细线）
1 & \checkmark & $\times$ & $\times$ & $\times$ & $\times$ 
  & \textbf{73.03} & \textbf{56.77} & \textbf{66.53} & \textbf{50.10} \\

2 & \checkmark & \checkmark & $\times$ & $\times$ & $\times$ 
  & 72.19 & 55.87 & 66.14 & 49.91 \\

3 & \checkmark & \checkmark & \checkmark & $\times$ & $\times$ 
  & 71.94 & 56.58 & 65.88 & 49.61 \\

4 & \checkmark & \checkmark & \checkmark & \checkmark & $\times$ 
  & 71.03 & 55.74 & 65.55 & 46.99 \\

5 & \checkmark & \checkmark & \checkmark & \checkmark & \checkmark 
  & 70.19 & 54.77 & 65.02 & 45.56 \\
\bottomrule % 标准学术底线（粗线）
\end{tabular}%
}
\vspace{2pt} 
\flushleft \footnotesize
\textit{Note:} FLED denotes the First Layer of Encoder and Decoder. Check marks indicate modules unfrozen during RL post-training.
\end{table}
\subsection{Ablation on Gradient-Isolation Strategy}

Additionally, we analyze the influence of different unfreezing strategies during RL post-training to experimentally validate the effectiveness of our gradient-isolation strategy. The results are shown in Table~\ref{tab:unfrozen_ablation}. In our default setting, as illustrated in Fig.~\ref{fig:post_training}, GIRL-DETR restricts RL training exclusively to the lightweight detection head ($\Theta_{\mathrm{head}}$) while keeping the multimodal backbone ($\Theta_{\mathrm{back}}$) strictly frozen. This configuration serves as our performance benchmark, achieving R1@0.5, R1@0.7, and mAP of 73.03\%, 56.77\%, and 50.10\%, respectively, and yielding a remarkable mIoU of 66.53\%.

The necessity of gradient isolation becomes increasingly evident as we systematically unfreeze other components. When the TGG module is unfrozen alongside the head, the R1@0.5, R1@0.7, and mAP performance metrics drop to 72.19\%, 55.87\%, and 49.91\%, respectively. This outcome suggests that exposing modules responsible for dynamic semantic injection to high-variance policy gradients weakens query awareness for temporal decision-making. Further unfreezing the CMI module causes mIoU and mAP to decrease to 65.88\% and 49.61\%, respectively, as updating the extensive parameter space governing multimodal distributions disrupts fragile cross-modal alignment and triggers representation instability. Additionally, we examine the effect of unfreezing the First Layer of Encoder and Decoder (FLED) as a transition, leading to a noticeable decline across all metrics, with mIoU falling to 65.55\% and mAP sliding to 46.99\%. Finally, when the entire network is fully unfrozen, the multimodal feature representations are further disrupted, resulting in continuous degradation across the board, with the mAP eventually dropping to 45.56\%.

These quantitative results demonstrate that while restricting RL updates to the detection head limits the optimization space, it effectively prevents the degradation of pre-trained feature manifolds. By keeping the fragile backbone frozen, GIRL-DETR achieves a crucial balance between training stability and direct metric optimization. Ultimately, this gradient-isolation strategy successfully alleviates the mismatch between surrogate losses and evaluation metrics, proving that gradient isolation is necessary to protect fragile modules when introducing RL into lightweight VMR architectures.

%%%下面的是我做的可视化实验，我已经写完了，你没必要写了，也不要改我部分

\enlargethispage{\baselineskip}
\section{Conclusion}

In this paper, we propose GIRL-DETR, a framework that introduces reinforcement learning post-training into lightweight VMR. This approach overcomes the optimization bottleneck caused by the discrepancy between continuous surrogate losses and the non-differentiable tIoU metric. By incorporating a text-guided gating mechanism to enhance query awareness and utilizing a gradient isolation strategy to freeze the feature backbone following supervised convergence, the proposed method optimizes solely the detection head. This structural design achieves an orthogonal decoupling between state representations and metric optimization. Experimental results across multiple benchmarks demonstrate that GIRL-DETR resolves the degradation of proxy losses with minimal parameter updates, thereby achieving superior localization accuracy and robust performance at exceptional efficiency.

Furthermore, the gradient-isolated design successfully prevents the degradation of pre-trained feature representations induced by high-variance policy gradients, while maintaining a negligible computational footprint during the reinforcement learning phase. As a pioneering attempt to introduce RL into lightweight VMR frameworks, this work establishes a highly robust and generalizable post-training paradigm. It demonstrates substantial alignment capacity without compromising established multimodal representations. Crucially, extensive evaluations demonstrate that the proposed post-training strategy can be effectively adapted to various representative baseline models for VMR. This adaptation yields consistent and significant performance gains, thereby validating the exceptional versatility and practical applicability of the strategy. In the future, we plan to extend this gradient-isolated alignment mechanism to broader multimodal spatio-temporal localization tasks, thereby further unlocking the efficiency and scalable potential of RL within resource-constrained architectures.

\nocite{*}
\bibliographystyle{IEEEtran}
\bibliography{Sections/6_references/references}

\begin{thebibliography}{10}
\providecommand{\url}[1]{#1}
\csname url@samestyle\endcsname
\providecommand{\newblock}{\relax}
\providecommand{\bibinfo}[2]{#2}
\providecommand{\BIBentrySTDinterwordspacing}{\spaceskip=0pt\relax}
\providecommand{\BIBentryALTinterwordstretchfactor}{4}
\providecommand{\BIBentryALTinterwordspacing}{\spaceskip=\fontdimen2\font plus
\BIBentryALTinterwordstretchfactor\fontdimen3\font minus \fontdimen4\font\relax}
\providecommand{\BIBforeignlanguage}[2]{{%
\expandafter\ifx\csname l@#1\endcsname\relax
\typeout{** WARNING: IEEEtran.bst: No hyphenation pattern has been}%
\typeout{** loaded for the language `#1'. Using the pattern for}%
\typeout{** the default language instead.}%
\else
\language=\csname l@#1\endcsname
\fi
#2}}
\providecommand{\BIBdecl}{\relax}
\BIBdecl

\bibitem{gao2017tall}
J.~Gao, C.~Sun, Z.~Yang, and R.~Nevatia, ``Tall: Temporal activity localization via language query,'' in \emph{Proc. IEEE Int. Conf. Comput. Vis. (ICCV)}, 2017, pp. 5267--5275.

\bibitem{anne2017localizing}
L.~Anne~Hendricks, O.~Wang, E.~Shechtman, J.~Sivic, T.~Darrell, and B.~Russell, ``Localizing moments in video with natural language,'' in \emph{Proc. IEEE Int. Conf. Comput. Vis. (ICCV)}, 2017, pp. 5803--5812.

\bibitem{sun2021maban}
X.~Sun, H.~Wang, and B.~He, ``Maban: Multi-agent boundary-aware network for natural language moment retrieval,'' \emph{IEEE Trans. Image Process.}, vol.~30, pp. 5589--5599, 2021.

\bibitem{zhang2023temporal}
H.~Zhang, A.~Sun, W.~Jing, and J.~T. Zhou, ``Temporal sentence grounding in videos: A survey and future directions,'' \emph{IEEE Trans. Pattern Anal. Mach. Intell.}, vol.~45, no.~8, pp. 10\,443--10\,465, 2023.

\bibitem{hu2021video}
Y.~Hu, M.~Liu, X.~Su, Z.~Gao, and L.~Nie, ``Video moment localization via deep cross-modal hashing,'' \emph{IEEE Trans. Image Process.}, vol.~30, pp. 4667--4677, 2021.

\bibitem{yang2021local}
W.~Yang, T.~Zhang, Y.~Zhang, and F.~Wu, ``Local correspondence network for weakly supervised temporal sentence grounding,'' \emph{IEEE Trans. Image Process.}, vol.~30, pp. 3252--3262, 2021.

\bibitem{yang2022video}
X.~Yang, S.~Wang, J.~Dong, J.~Dong, M.~Wang, and T.-S. Chua, ``Video moment retrieval with cross-modal neural architecture search,'' \emph{IEEE Trans. Image Process.}, vol.~31, pp. 1204--1216, 2022.

\bibitem{lee2024bam}
P.~Lee and H.~Byun, ``Bam-detr: Boundary-aligned moment detection transformer for temporal sentence grounding in videos,'' in \emph{Proc. Eur. Conf. Comput. Vis. (ECCV)}.\hskip 1em plus 0.5em minus 0.4em\relax Springer, 2024, pp. 220--238.

\bibitem{seol2023bmrn}
M.~Seol, J.~Kim, and J.~Moon, ``Bmrn: Boundary matching and refinement network for temporal moment localization with natural language,'' in \emph{Proc. IEEE/CVF Conf. Comput. Vis. Pattern Recognit. (CVPR)}, 2023, pp. 5571--5579.

\bibitem{wang2026time}
Y.~Wang, Z.~Wang, B.~Xu, Y.~Du, K.~Lin, Z.~Xiao, Z.~Yue, J.~Ju, L.~Zhang, D.~Yang \emph{et~al.}, ``Time-r1: Post-training large vision language model for temporal video grounding,'' \emph{Adv. Neural Inf. Process. Syst. (NIPS)}, vol.~38, pp. 83\,330--83\,364, 2026.

\bibitem{chen2025datasets}
R.~Chen, T.~Luo, Z.~Fan, H.~Zou, Z.~Feng, G.~Xie, H.~Zhang, Z.~Wang, Z.~Liu, and Z.~Huaijian, ``Datasets and recipes for video temporal grounding via reinforcement learning,'' in \emph{Proc. Conf. Empir. Methods Nat. Lang. Process. (EMNLP)}, 2025, pp. 983--992.

\bibitem{lei2021detecting}
J.~Lei, T.~L. Berg, and M.~Bansal, ``Detecting moments and highlights in videos via natural language queries,'' \emph{Adv. Neural Inf. Process. Syst. (NIPS)}, vol.~34, pp. 11\,846--11\,858, 2021.

\bibitem{Moon2025CorrelationguidedCO}
\BIBentryALTinterwordspacing
W.~Moon, S.~Hyun, S.~Lee, and J.~pil Heo, ``Correlation-guided calibration of query dependency for video temporal grounding,'' \emph{Pattern Recognit.}, vol. 174, p. 112984, 2025. [Online]. Available: \url{https://api.semanticscholar.org/CorpusID:284322691}
\BIBentrySTDinterwordspacing

\bibitem{sun2024tr}
H.~Sun, M.~Zhou, W.~Chen, and W.~Xie, ``Tr-detr: Task-reciprocal transformer for joint moment retrieval and highlight detection,'' in \emph{Proc. {AAAI} Conf. Artif. Intell.}, vol.~38, no.~5, 2024, pp. 4998--5007.

\bibitem{liu2022umt}
Y.~Liu, S.~Li, Y.~Wu, C.-W. Chen, Y.~Shan, and X.~Qie, ``Umt: Unified multi-modal transformers for joint video moment retrieval and highlight detection,'' in \emph{Proc. IEEE/CVF Conf. Comput. Vis. Pattern Recognit. (CVPR)}, 2022, pp. 3042--3051.

\bibitem{xiao2024bridging}
Y.~Xiao, Z.~Luo, Y.~Liu, Y.~Ma, H.~Bian, Y.~Ji, Y.~Yang, and X.~Li, ``Bridging the gap: A unified video comprehension framework for moment retrieval and highlight detection,'' in \emph{Proc. IEEE/CVF Conf. Comput. Vis. Pattern Recognit. (CVPR)}, 2024, pp. 18\,709--18\,719.

\bibitem{ouyang2022training}
L.~Ouyang, J.~Wu, X.~Jiang, D.~Almeida, C.~Wainwright, P.~Mishkin, C.~Zhang, S.~Agarwal, K.~Slama, A.~Ray \emph{et~al.}, ``Training language models to follow instructions with human feedback,'' \emph{Adv. Neural Inf. Process. Syst. (NIPS)}, vol.~35, pp. 27\,730--27\,744, 2022.

\bibitem{sun2024aligning}
Z.~Sun, S.~Shen, S.~Cao, H.~Liu, C.~Li, Y.~Shen, C.~Gan, L.~Gui, Y.-X. Wang, Y.~Yang \emph{et~al.}, ``Aligning large multimodal models with factually augmented rlhf,'' in \emph{Findings Assoc. Comput. Linguist. (ACL)}, 2024, pp. 13\,088--13\,110.

\bibitem{feng2026video}
K.~Feng, K.~Gong, B.~Li, Z.~Guo, Y.~Wang, T.~Peng, J.~Wu, X.~Zhang, B.~Wang, and X.~Yue, ``Video-r1: Reinforcing video reasoning in mllms,'' \emph{Adv. Neural Inf. Process. Syst. (NIPS)}, vol.~38, pp. 99\,114--99\,137, 2026.

\bibitem{williams1992simple}
R.~J. Williams, ``Simple statistical gradient-following algorithms for connectionist reinforcement learning,'' \emph{Mach. Learn.}, vol.~8, no.~3, pp. 229--256, 1992.

\bibitem{sutton1999policy}
R.~S. Sutton, D.~McAllester, S.~Singh, and Y.~Mansour, ``Policy gradient methods for reinforcement learning with function approximation,'' \emph{Adv. Neural Inf. Process. Syst. (NIPS)}, vol.~12, 1999.

\bibitem{kirkpatrick2017overcoming}
J.~Kirkpatrick, R.~Pascanu, N.~Rabinowitz, J.~Veness, G.~Desjardins, A.~A. Rusu, K.~Milan, J.~Quan, T.~Ramalho, A.~Grabska-Barwinska \emph{et~al.}, ``Overcoming catastrophic forgetting in neural networks,'' \emph{Proc. Natl. Acad. Sci.}, vol. 114, no.~13, pp. 3521--3526, 2017.

\bibitem{li2017learning}
Z.~Li and D.~Hoiem, ``Learning without forgetting,'' \emph{IEEE Trans. Pattern Anal. Mach. Intell.}, vol.~40, no.~12, pp. 2935--2947, 2017.

\bibitem{liu2021context}
D.~Liu, X.~Qu, J.~Dong, P.~Zhou, Y.~Cheng, W.~Wei, Z.~Xu, and Y.~Xie, ``Context-aware biaffine localizing network for temporal sentence grounding,'' in \emph{Proc. IEEE/CVF Conf. Comput. Vis. Pattern Recognit. (CVPR)}, 2021, pp. 11\,235--11\,244.

\bibitem{regneri2013grounding}
M.~Regneri, M.~Rohrbach, D.~Wetzel, S.~Thater, B.~Schiele, and M.~Pinkal, ``Grounding action descriptions in videos,'' \emph{Trans. Assoc. Comput. Linguist.}, vol.~1, pp. 25--36, 2013.

\bibitem{yuan2019find}
Y.~Yuan, T.~Mei, and W.~Zhu, ``To find where you talk: Temporal sentence localization in video with attention based location regression,'' in \emph{Proc. {AAAI} Conf. Artif. Intell.}, vol.~33, no.~01, 2019, pp. 9159--9166.

\bibitem{zhang2020learning}
S.~Zhang, H.~Peng, J.~Fu, and J.~Luo, ``Learning 2d temporal adjacent networks for moment localization with natural language,'' in \emph{Proc. {AAAI} Conf. Artif. Intell.}, vol.~34, no.~07, 2020, pp. 12\,870--12\,877.

\bibitem{zhang2020span}
H.~Zhang, A.~Sun, W.~Jing, and J.~T. Zhou, ``Span-based localizing network for natural language video localization,'' in \emph{Proc. 58th Annu. Meet. Assoc. Comput. Linguist. (ACL)}, 2020, pp. 6543--6554.

\bibitem{zeng2020dense}
R.~Zeng, H.~Xu, W.~Huang, P.~Chen, M.~Tan, and C.~Gan, ``Dense regression network for video grounding,'' in \emph{Proc. IEEE/CVF Conf. Comput. Vis. Pattern Recognit. (CVPR)}, 2020, pp. 10\,287--10\,296.

\bibitem{chen2020learning}
S.~Chen, W.~Jiang, W.~Liu, and Y.-G. Jiang, ``Learning modality interaction for temporal sentence localization and event captioning in videos,'' in \emph{Proc. Eur. Conf. Comput. Vis. (ECCV)}.\hskip 1em plus 0.5em minus 0.4em\relax Springer, 2020, pp. 333--351.

\bibitem{zhang2021multi}
Z.~Zhang, X.~Han, X.~Song, Y.~Yan, and L.~Nie, ``Multi-modal interaction graph convolutional network for temporal language localization in videos,'' \emph{IEEE Trans. Image Process.}, vol.~30, pp. 8265--8277, 2021.

\bibitem{cao2021pursuit}
M.~Cao, L.~Chen, M.~Z. Shou, C.~Zhang, and Y.~Zou, ``On pursuit of designing multi-modal transformer for video grounding,'' in \emph{Proc. Conf. Empir. Methods Nat. Lang. Process. (EMNLP)}, 2021, pp. 9810--9823.

\bibitem{moon2023query}
W.~Moon, S.~Hyun, S.~Park, D.~Park, and J.-P. Heo, ``Query-dependent video representation for moment retrieval and highlight detection,'' in \emph{Proc. IEEE/CVF Conf. Comput. Vis. Pattern Recognit. (CVPR)}, 2023, pp. 23\,023--23\,033.

\bibitem{jang2023knowing}
J.~Jang, J.~Park, J.~Kim, H.~Kwon, and K.~Sohn, ``Knowing where to focus: Event-aware transformer for video grounding,'' in \emph{Proc. IEEE Int. Conf. Comput. Vis. (ICCV)}, 2023, pp. 13\,846--13\,856.

\bibitem{christiano2017deep}
P.~F. Christiano, J.~Leike, T.~Brown, M.~Martic, S.~Legg, and D.~Amodei, ``Deep reinforcement learning from human preferences,'' \emph{Adv. Neural Inf. Process. Syst.}, vol.~30, 2017.

\bibitem{stiennon2020learning}
N.~Stiennon, L.~Ouyang, J.~Wu, D.~Ziegler, R.~Lowe, C.~Voss, A.~Radford, D.~Amodei, and P.~F. Christiano, ``Learning to summarize with human feedback,'' \emph{Adv. Neural Inf. Process. Syst.}, vol.~33, pp. 3008--3021, 2020.

\bibitem{touvron2023llama}
H.~Touvron, L.~Martin, K.~Stone, P.~Albert, A.~Almahairi, Y.~Babaei, N.~Bashlykov, S.~Batra, P.~Bhargava, S.~Bhosale \emph{et~al.}, ``Llama 2: Open foundation and fine-tuned chat models,'' \emph{arXiv preprint arXiv:2307.09288}, 2023.

\bibitem{rafailov2023direct}
R.~Rafailov, A.~Sharma, E.~Mitchell, C.~D. Manning, S.~Ermon, and C.~Finn, ``Direct preference optimization: Your language model is secretly a reward model,'' \emph{Adv. Neural Inf. Process. Syst. (NIPS)}, vol.~36, pp. 53\,728--53\,741, 2023.

\bibitem{dettmers2023qlora}
T.~Dettmers, A.~Pagnoni, A.~Holtzman, and L.~Zettlemoyer, ``Qlora: Efficient finetuning of quantized llms,'' \emph{Adv. Neural Inf. Process. Syst.}, vol.~36, pp. 10\,088--10\,115, 2023.

\bibitem{hu2022lora}
E.~J. Hu, Y.~Shen, P.~Wallis, Z.~Allen-Zhu, Y.~Li, S.~Wang, L.~Wang, W.~Chen \emph{et~al.}, ``Lora: Low-rank adaptation of large language models.'' \emph{Proc. 10th Int. Conf. Learn. Represent. (ICLR)}, 2022.

\bibitem{houlsby2019parameter}
N.~Houlsby, A.~Giurgiu, S.~Jastrzebski, B.~Morrone, Q.~De~Laroussilhe, A.~Gesmundo, M.~Attariyan, and S.~Gelly, ``Parameter-efficient transfer learning for nlp,'' in \emph{Proc. Int. Conf. Mach. Learn. (ICML)}.\hskip 1em plus 0.5em minus 0.4em\relax PMLR, 2019, pp. 2790--2799.

\bibitem{caicedo2015active}
J.~C. Caicedo and S.~Lazebnik, ``Active object localization with deep reinforcement learning,'' in \emph{Proc. IEEE Int. Conf. Comput. Vis. (ICCV)}, 2015, pp. 2488--2496.

\bibitem{he2019read}
D.~He, X.~Zhao, J.~Huang, F.~Li, X.~Liu, and S.~Wen, ``Read, watch, and move: Reinforcement learning for temporally grounding natural language descriptions in videos,'' in \emph{Proc. {AAAI} Conf. Artif. Intell.}, vol.~33, no.~01, 2019, pp. 8393--8400.

\bibitem{wu2020tree}
J.~Wu, G.~Li, S.~Liu, and L.~Lin, ``Tree-structured policy based progressive reinforcement learning for temporally language grounding in video,'' in \emph{Proc. {AAAI} Conf. Artif. Intell.}, vol.~34, no.~07, 2020, pp. 12\,386--12\,393.

\bibitem{shao2024deepseekmath}
Z.~Shao, P.~Wang, Q.~Zhu, R.~Xu, J.~Song, X.~Bi, H.~Zhang, M.~Zhang, Y.~Li, Y.~Wu \emph{et~al.}, ``Deepseekmath: Pushing the limits of mathematical reasoning in open language models,'' \emph{arXiv preprint arXiv:2402.03300}, 2024.

\bibitem{wang2023spatiotemporal}
Y.~Wang, K.~Li, G.~Chen, Y.~Zhang, D.~Guo, and M.~Wang, ``Spatiotemporal contrastive modeling for video moment retrieval,'' \emph{World Wide Web}, vol.~26, no.~4, pp. 1525--1544, 2023.

\bibitem{li2023momentdiff}
P.~Li, C.-W. Xie, H.~Xie, L.~Zhao, L.~Zhang, Y.~Zheng, D.~Zhao, and Y.~Zhang, ``Momentdiff: Generative video moment retrieval from random to real,'' \emph{Adv. Neural Inf. Process. Syst. (NIPS)}, vol.~36, pp. 65\,948--65\,966, 2023.

\bibitem{lin2023univtg}
K.~Q. Lin, P.~Zhang, J.~Chen, S.~Pramanick, D.~Gao, A.~J. Wang, R.~Yan, and M.~Z. Shou, ``Univtg: Towards unified video-language temporal grounding,'' in \emph{Proc. IEEE Int. Conf. Comput. Vis. (ICCV)}, 2023, pp. 2794--2804.

\bibitem{jiang2024prior}
Y.~Jiang, W.~Zhang, X.~Zhang, X.-Y. Wei, C.~W. Chen, and Q.~Li, ``Prior knowledge integration via llm encoding and pseudo event regulation for video moment retrieval,'' in \emph{Proc. 32nd ACM Int. Conf. Multimedia}, 2024, pp. 7249--7258.

\bibitem{sun2025diversifying}
X.~Sun, L.~Shi, L.~Wang, S.~Zhou, K.~Xia, Y.~Wang, and G.~Hua, ``Diversifying query: Region-guided transformer for temporal sentence grounding,'' in \emph{Proc. {AAAI} Conf. Artif. Intell.}, vol.~39, no.~7, 2025, pp. 7131--7139.

\bibitem{hu2024maskable}
J.~Hu, D.~Guo, K.~Li, Z.~Si, X.~Yang, and M.~Wang, ``Maskable retentive network for video moment retrieval,'' in \emph{Proc. 32nd ACM Int. Conf. Multimedia}, 2024, pp. 1476--1485.

\bibitem{xiong2025sa}
T.~Xiong, W.~Wei, K.~Xu, and D.~Chen, ``Sa-detr: Span aware detection transformer for moment retrieval,'' in \emph{Proc. 31st Int. Conf. Comput. Linguist. (COLING)}, 2025, pp. 7634--7647.

\bibitem{hao2022query}
J.~Hao, H.~Sun, P.~Ren, J.~Wang, Q.~Qi, and J.~Liao, ``Query-aware video encoder for video moment retrieval,'' \emph{Neurocomputing}, vol. 483, pp. 72--86, 2022.

\bibitem{dosovitskiy2020image}
A.~Dosovitskiy, L.~Beyer, A.~Kolesnikov, D.~Weissenborn, X.~Zhai, T.~Unterthiner, M.~Dehghani, M.~Minderer, G.~Heigold, S.~Gelly \emph{et~al.}, ``An image is worth 16x16 words: Transformers for image recognition at scale,'' \emph{arXiv preprint arXiv:2010.11929}, 2020.

\bibitem{radford2021learning}
A.~Radford, J.~W. Kim, C.~Hallacy, A.~Ramesh, G.~Goh, S.~Agarwal, G.~Sastry, A.~Askell, P.~Mishkin, J.~Clark \emph{et~al.}, ``Learning transferable visual models from natural language supervision,'' in \emph{Proc. Int. Conf. Mach. Learn. (ICML)}.\hskip 1em plus 0.5em minus 0.4em\relax PMLR, 2021, pp. 8748--8763.

\bibitem{li2022blip}
J.~Li, D.~Li, C.~Xiong, and S.~Hoi, ``Blip: Bootstrapping language-image pre-training for unified vision-language understanding and generation,'' in \emph{Proc. Int. Conf. Mach. Learn. (ICML)}.\hskip 1em plus 0.5em minus 0.4em\relax PMLR, 2022, pp. 12\,888--12\,900.

\bibitem{wang2024internvideo2}
Y.~Wang, K.~Li, X.~Li, J.~Yu, Y.~He, G.~Chen, B.~Pei, R.~Zheng, J.~Xu, Z.~Wang \emph{et~al.}, ``Internvideo2: Scaling video foundation models for multimodal video understanding. arxiv 2024,'' \emph{arXiv preprint arXiv:2403.15377}, vol.~2, 2024.

\bibitem{feichtenhofer2019slowfast}
C.~Feichtenhofer, H.~Fan, J.~Malik, and K.~He, ``Slowfast networks for video recognition,'' in \emph{Proc. IEEE Int. Conf. Comput. Vis. (ICCV)}, 2019, pp. 6202--6211.

\bibitem{paul2024videolights}
D.~Paul, M.~R. Parvez, N.~Mohammed, and S.~Rahman, ``Videolights: Feature refinement and cross-task alignment transformer for joint video highlight detection and moment retrieval,'' \emph{arXiv preprint arXiv:2412.01558}, 2024.

\end{thebibliography}

\begin{IEEEbiography}[{\includegraphics[width=1in,height=1.25in,clip,keepaspectratio]{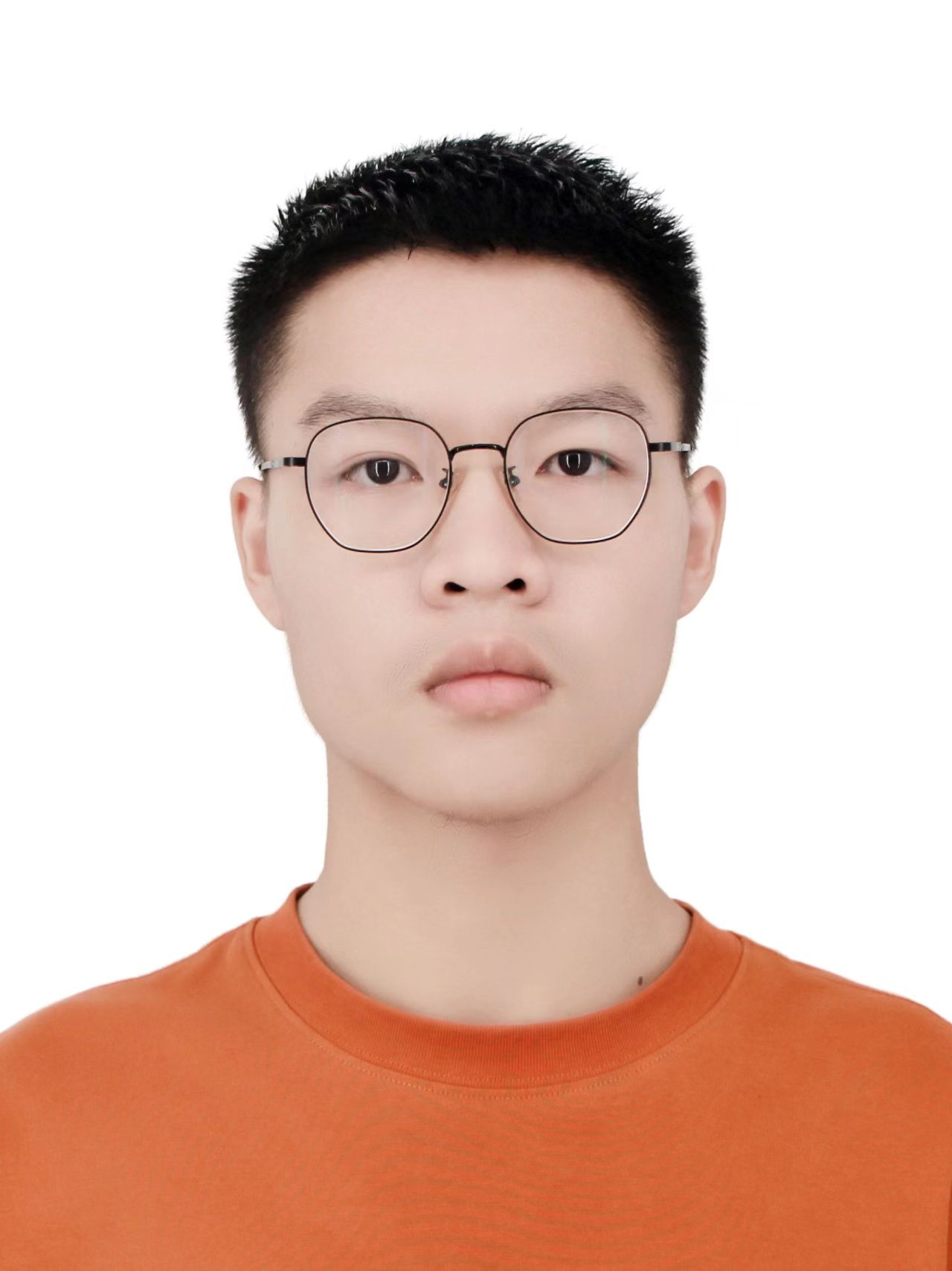}}]{Shihang Zhang} is currently pursuing the B.E. degree with Sichuan University. His research interests include computer vision and video understanding with multi-modal large models, with a particular focus on object detection, attribute recognition, and video temporal localization.
\end{IEEEbiography}

\begin{IEEEbiography}[{\includegraphics[width=1in,height=1.25in,clip,keepaspectratio]{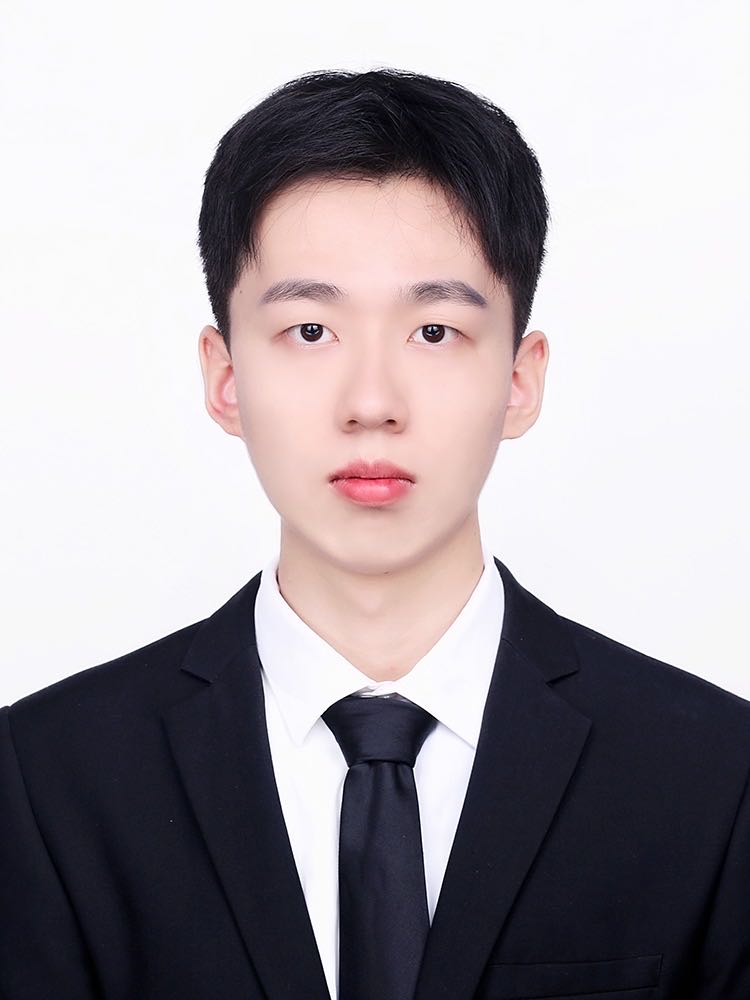}}]{Mingjin Kuai} received his M.Eng. degree from Central South University in 2025. He is currently pursuing a PhD at Nanjing University. He has been invited to serve as a reviewer for journals such as ACM TOMM, IEEE TCSVT, and IJCV. His research interests include multi-modal analysis and retrieval, and multi-modal large models.
\end{IEEEbiography}

\begin{IEEEbiography}[{\includegraphics[width=1in,height=1.25in,clip,keepaspectratio]{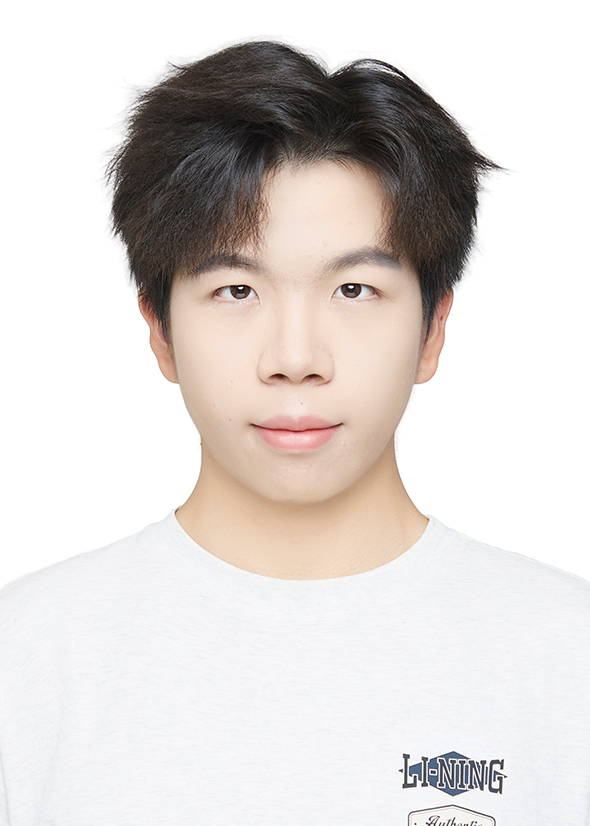}}]{Ye Wei} is currently pursuing the B.E. degree with Sichuan University. His research interests include test-time adaptation and video understanding with multi-modal large models, with a particular focus on text-video retrieval, visual grounding, and video temporal localization.
\end{IEEEbiography}

\vfill
\begin{IEEEbiography}[{\includegraphics[width=1in,height=1.25in,clip,keepaspectratio]{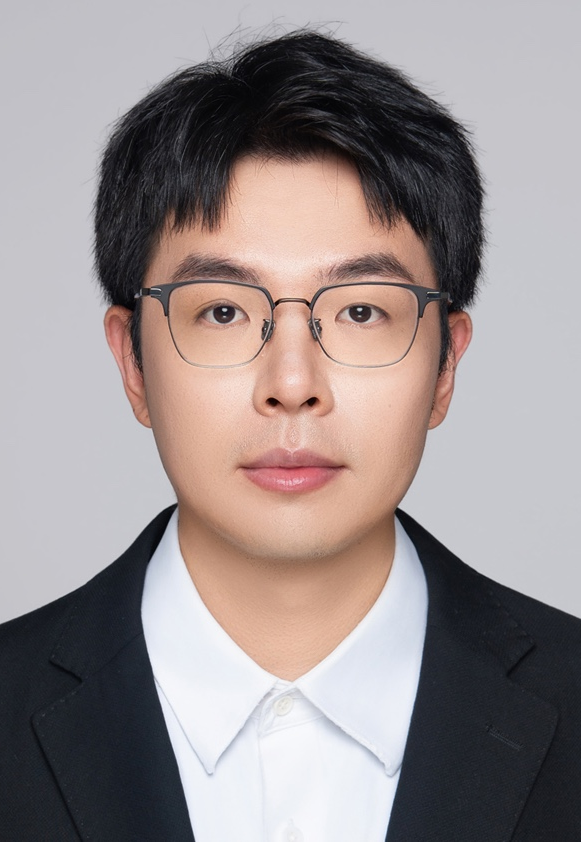}}]{Zhen Zhang}
received the PhD degree in computer science from Zhejiang University, China, in 2021. He is currently an Assistant Professor in Nanjing University. His research interests include machine learning, data mining and artificial intelligence. He has published more than 30 papers in refereed journals and conferences.
\end{IEEEbiography}

\begin{IEEEbiography}[{\includegraphics[width=1in,height=1.25in,clip,keepaspectratio]{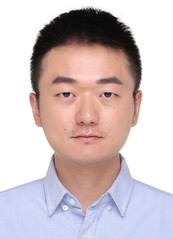}}]{Wei Ji}
is an associate professor at Nanjing University. He was a Research Fellow in the School of Computing at National University of Singapore. He received the Ph.D. degree in computer science from the Zhejiang University in 2020. He has published more than 80 papers in top conferences such as CVPR, ECCV, SIGIR, AAAI, and journals including TPAMI, TIP and TCYB.  He serves as Associate Editor of IEEE TIP, IEEE TCSVT, PR and ACM TOMM and Area Chair of NeurIPS, IJCAI, ACM Multimedia and so on. His current research interests include multi-modal learning, and cross-modal retrieval.
\end{IEEEbiography}

\vfill
\end{document}